\documentclass[10pt,journal,compsoc]{IEEEtran}

\ifCLASSOPTIONcompsoc
  \usepackage[nocompress]{cite}
\else
  \usepackage{cite}
\fi 

\usepackage{times}
\usepackage{epsfig} 
\usepackage{amssymb}
\usepackage{hyperref}

\hypersetup{
    colorlinks=true,
    linkcolor=red,
    citecolor=cyan,
    filecolor=magenta,      
    urlcolor=cyan,
    }

\usepackage{enumerate} 
\usepackage{amsthm}
\usepackage{color}
\usepackage{algpseudocode} 
\usepackage{pifont}
 \newcommand{\cmark}{\ding{51}}%
\include{def} 
\usepackage{tabu} 
   \usepackage{float}
\usepackage{subfigure}
\usepackage{wrapfig}
\usepackage[table]{xcolor}
\usepackage{microtype}
\usepackage{graphicx}
\usepackage{subfigure} 
\usepackage{bm}
\usepackage{comment}
\usepackage{amsfonts}
\usepackage{amsmath}
\usepackage{balance} 
\usepackage{listings} 
\usepackage{xcolor} 
\usepackage{arydshln}
\usepackage{cases} 
\usepackage{booktabs}
\usepackage{algorithm}
\usepackage{multicol,multirow}   
\usepackage{threeparttable} 
\def\ie{\emph{i.e., }}
\def\eg{\emph{e.g., }}

 \usepackage{tikz}
\newcommand{\tikzxmark}{%
\tikz[scale=0.23] {
    \draw[line width=0.7,line cap=round] (0,0) to [bend left=6] (1,1);
    \draw[line width=0.7,line cap=round] (0.2,0.95) to [bend right=3] (0.8,0.05);
}}
\newcommand{\tikzcmark}{%
\tikz[scale=0.23] {
    \draw[line width=0.7,line cap=round] (0.25,0) to [bend left=10] (1,1);
    \draw[line width=0.8,line cap=round] (0,0.35) to [bend right=1] (0.23,0);
}}

\begin{document}
 
\title{Deep Long-Tailed  Learning: A Survey}
 
\author{Yifan Zhang, Bingyi Kang, Bryan Hooi, Shuicheng Yan,~\IEEEmembership{Fellow,~IEEE}, and Jiashi Feng 
\IEEEcompsocitemizethanks{\IEEEcompsocthanksitem Y. Zhang and B. Hooi are with School of Computing, National University of Singapore. E-mail: yifan.zhang@u.nus.edu, dcsbhk@nus.edu.sg. 
\IEEEcompsocthanksitem B. Kang and J. Feng are with ByteDance AI Lab. E-mail: bingykang@gmail.com,  jshfeng@bytedance.com.  
\IEEEcompsocthanksitem S. Yan is with SEA AI Lab. E-mail: yansc@sea.com. 
}
\thanks{}} 

%
%


\markboth{IEEE Transactions on Pattern Analysis and Machine Intelligence}%
{Shell \MakeLowercase{\textit{et al.}}: Bare Advanced Demo of IEEEtran.cls for IEEE Computer Society Journals}

%

\IEEEtitleabstractindextext{%
\begin{abstract} 
Deep long-tailed learning, one of the most challenging problems in visual recognition, aims to train well-performing deep models from a large number of images that follow a long-tailed class distribution. In the last decade, deep learning has emerged as a powerful recognition model for learning high-quality image representations and has led to remarkable breakthroughs in generic visual recognition. However, long-tailed class imbalance, a common problem in practical visual recognition tasks, often limits the practicality of deep network based recognition models in real-world applications, since they can be easily biased towards dominant classes and perform poorly on tail classes.  To address this problem, a large number of studies have been conducted in recent years, making promising progress in the field of deep long-tailed learning. Considering the rapid evolution of this field,  this paper aims to provide a comprehensive survey on recent advances in deep long-tailed learning. To be specific, we group existing deep long-tailed learning studies into three main categories (\ie class re-balancing, information augmentation and module improvement), and review these methods  following this taxonomy in detail. Afterward, we empirically analyze several state-of-the-art  methods by evaluating to what extent they   address the issue of class imbalance via a newly proposed evaluation metric, \ie relative accuracy.  We conclude the survey by highlighting important applications of deep long-tailed learning and identifying several promising directions for future research.

\end{abstract}

}

\maketitle

\IEEEdisplaynontitleabstractindextext 

%
\IEEEpeerreviewmaketitle

\ifCLASSOPTIONcompsoc
\IEEEraisesectionheading{\section{Introduction}\label{sec:introduction}}
\else
\section{Introduction}
\label{sec:introduction}
\fi
 
\IEEEPARstart{D}{eep}   learning   allows computational models, composed of multiple processing layers, to learn data representations with multiple levels of abstraction~\cite{lecun2015deep,goodfellow2016deep} and has made incredible progress in computer vision~\cite{voulodimos2018deep,dong2015image,wang2020deep,krizhevsky2012imagenet,ren2015faster,shelhamer2016fully}. The key enablers of deep learning are  the availability of large-scale datasets, the emergence  of GPUs, and the advancement of deep network architectures~\cite{bengio2020deep}. Thanks to the strong ability of learning high-quality data representations, deep neural networks have been applied with great success to many visual discriminative tasks, including   image classification~\cite{krizhevsky2012imagenet,he2016deep}, object detection~\cite{szegedy2013deep,ren2015faster} and semantic segmentation~\cite{girshick2014rich,shelhamer2016fully}.

\begin{figure}[t]
 \centering
  \includegraphics[width=8.5cm]{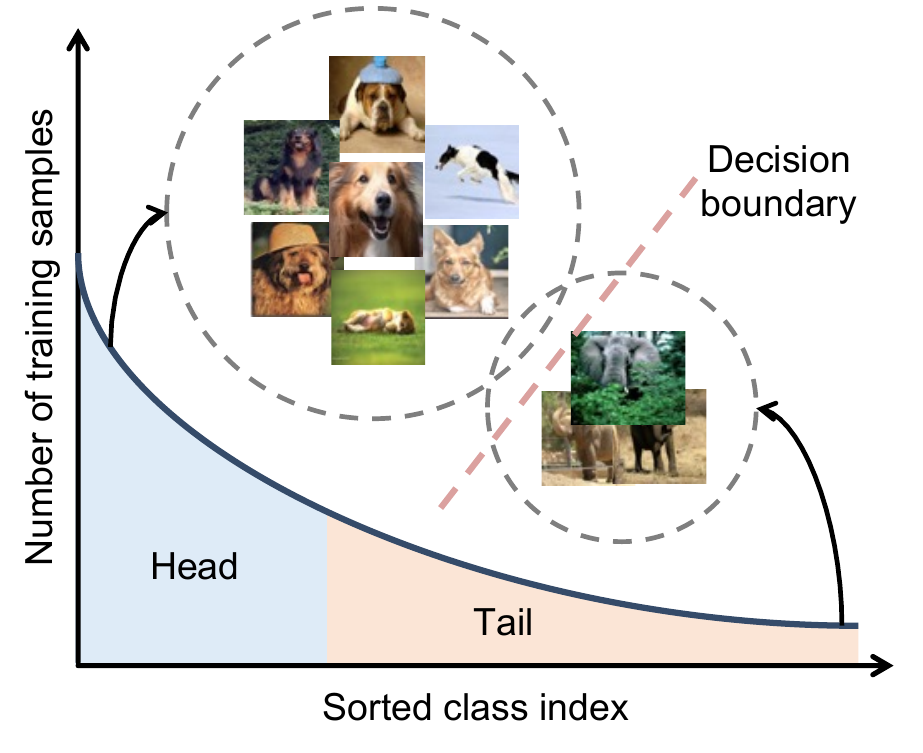}
 \vspace{-0.1in}
  \caption{The label distribution of a long-tailed   dataset (\eg the iNaturalist species dataset~\cite{van2018inaturalist} with more than 8,000 classes). The head-class  feature space    learned  on these sampled is often larger than tail classes, while the decision boundary is usually biased towards dominant classes.}   
  \label{fig_long_tail} \vspace{-0.1in}
\end{figure}

In real-world  applications,   training samples  typically exhibit a long-tailed  class distribution, where a small portion of classes have {a massive number of}  sample points but the others are associated with only a few samples~\cite{kang2021exploring,menon2020long,liu2019large,cui2019class}. Such class imbalance of training sample numbers, however, makes the training of  deep  network based recognition models  very challenging. As shown in Fig.~\ref{fig_long_tail}, the trained model can be easily biased towards    head  classes with massive training data, leading to poor model performance on tail  classes that   have    limited data~\cite{wang2020long,cao2019learning,tan2020equalization}. Therefore,  the deep models trained by the common practice of  empirical risk minimization~\cite{vapnik1992principles} cannot  handle real-world applications with long-tailed class imbalance, \eg  face recognition~\cite{zhang2017range,cao2020domain},  species classification~\cite{van2018inaturalist,miao2021iterative},  medical image diagnosis~\cite{ju2021relational}, urban scene understanding~\cite{he2021re} and unmanned aerial vehicle detection~\cite{yu2021towards}.

To address long-tailed class imbalance, massive deep long-tailed learning studies have been conducted     in recent years~\cite{cui2019class,jamal2020rethinking,liu2019large,zhang2021distribution,zhang2021test}. Despite the rapid evolution in this field, there is still no systematic study to review and discuss existing progress.
To fill this gap, we aim  to provide a comprehensive  survey for recent long-tailed learning studies conducted before mid-2021. 

\begin{figure}[t]
  \includegraphics[width=8.8cm]{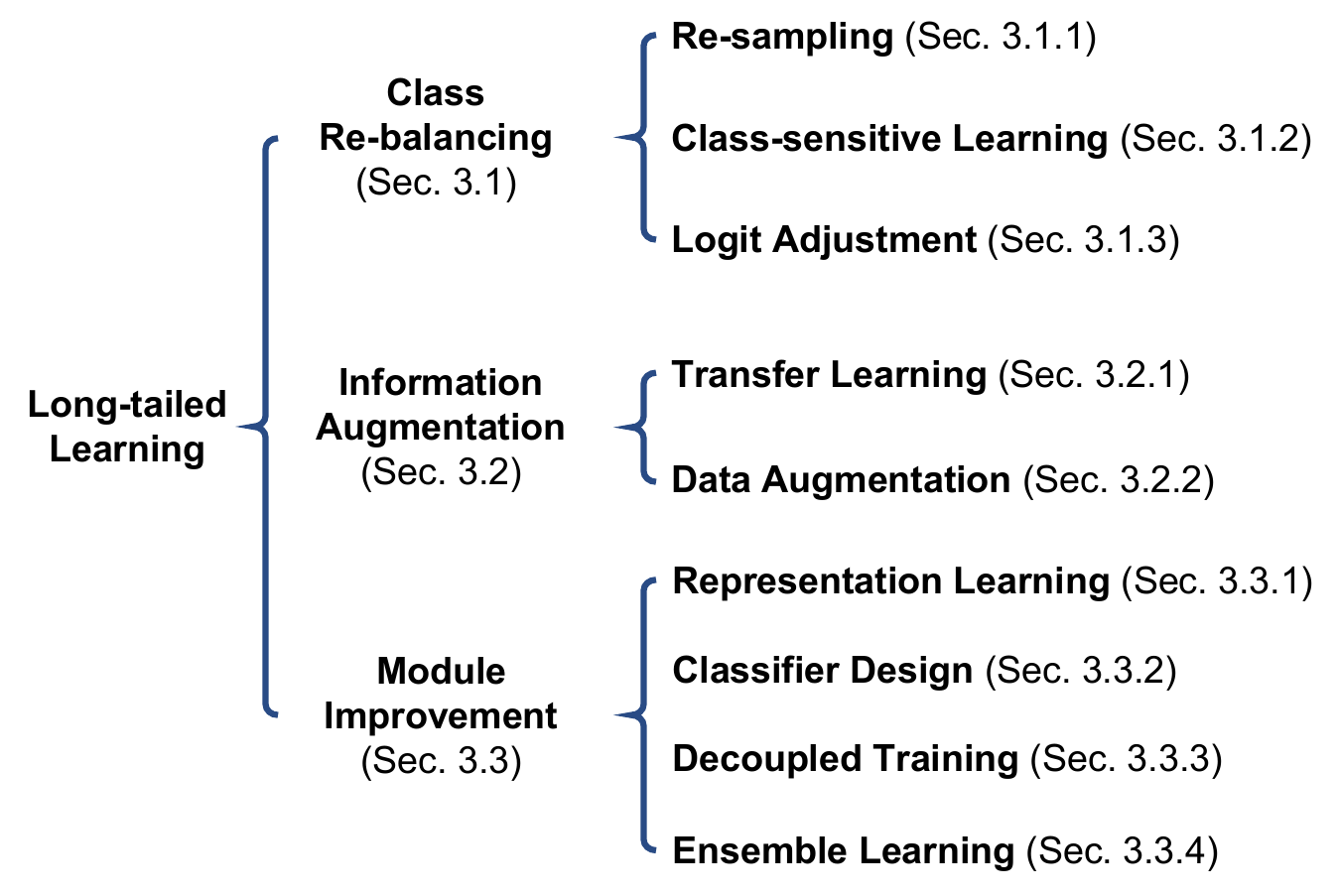}
  \vspace{-0.1in}
  \caption{Taxonomy of existing deep long-tailed learning methods.}   
  \label{fig_taxonomy}   \vspace{-0.1in}
\end{figure}

As shown in Fig.~\ref{fig_taxonomy}, we group existing  methods into three main categories based on their main technical contributions, \ie class re-balancing, information augmentation and module improvement; these categories can be further  classified into nine sub-categories: re-sampling, class-sensitive learning, logit adjustment, transfer learning, data augmentation,   representation learning, classifier design, decoupled training and ensemble learning. According to this taxonomy, we   provide a  comprehensive review of existing methods, and also empirically  analyze several state-of-the-art methods by evaluating their abilities of handling class imbalance using a new  evaluation metric, namely \textit{relative accuracy}. We conclude the survey by introducing several real-world  application scenarios of deep long-tailed learning and identifying several promising research directions that can be explored by the community in the future.

We summarize the key contributions of this survey as follows.
\begin{itemize}
    \item To the best of our   knowledge, this is the first comprehensive survey of deep long-tailed learning, which will  provide a better understanding of long-tailed visual learning with deep neural networks  for researchers and the community.
    
    \item We provide an in-depth review of  advanced long-tailed learning studies, and empirically  study   state-of-the-art methods by evaluating to what extent they   handle long-tailed  class imbalance via a new relative accuracy metric.
  
    \item We identify four potential directions for method innovation as well as eight new deep long-tailed learning task settings for future research. 
   
\end{itemize}

The rest of this survey will be organized as follows:  Section~\ref{Sec2} presents the problem definition and introduces widely-used datasets,  metrics and  applications. Section~\ref{sec_method} provides a comprehensive review of advanced  long-tailed learning methods and Section~\ref{Sec4} empirically analyzes several state-of-the-art  methods based on a new   evaluation metric.   Section~\ref{Sec6} identifies future research directions. We conclude the survey in Section~\ref{Sec7}.

\section{{Problem Overview}}\label{Sec2}

\subsection{Problem Definition}
Deep long-tailed learning seeks to learn a deep neural network model from a training dataset  with a long-tailed class distribution, where a small fraction of classes have {a massive number of samples, and the rest of the  classes} are associated with only a few samples (c.f. Fig.~\ref{fig_long_tail}). Let $\{x_i,y_i\}_{i=1}^{n}$ be the long-tailed training  set, where each sample $x_i$ has a corresponding class label $y_i$. The total number of training set over $K$ classes is $n = \sum_{k=1}^K n_k$, where  $n_k$ denotes the data number of  class $k$; let
$\pi$ denote the vector of label frequencies, where  $\pi_k =n_k/n$ indicates the label frequency of  class $k$. 
Without loss of generality, a common assumption in long-tailed learning~\cite{hong2020disentangling,kang2019decoupling} is that the  classes are sorted by cardinality in decreasing order (\ie if $i_1 < i_2$, then $n_{i_1}\geq n_{i_2}$, and $n_1 \gg n_K$), and then the imbalance ratio is defined as $n_1$/$n_K$. 

This task is challenging   due to two  difficulties: (1)   imbalanced   data numbers across classes make   deep models      biased to   head   classes and performs poorly on tail  classes; (2) lack of tail-class samples makes it further challenging to train models for  tail-class classification. Such a task is fundamental and may occur in various visual recognition tasks, such as image classification~\cite{liu2019large,kang2019decoupling}, detection~\cite{tan2020equalization,feng2021exploring } and segmentation~\cite{wang2020devil,weng2021unsupervised,he2021re}.


\begin{table}[h]    
    \caption{Statistics of long-tailed   datasets.   ``Cls." indicates image classification; ``Det." represents object detection; ``Seg." means instance segmentation.}\label{dataset}  
    \begin{center}   \vspace{-0.1in}
    \scalebox{0.8}{  
    \begin{threeparttable} 
	\begin{tabular}{llccc}\toprule
      Task  &Dataset & $\#$ classes &  $\#$ training data & $\#$ test data    \\ \midrule    
     \multirow{4}{*}{Cls.}   &ImageNet-LT~\cite{liu2019large} 	& 1,000 & 115,846 &50,000 	  \\ 
        &CIFAR100-LT~\cite{cao2019learning} & 100 & 50,000 & 10,000   \\
        &Places-LT~\cite{liu2019large} & 365 & 62,500 & 36,500   \\
        &iNaturalist 2018~\cite{van2018inaturalist} & 8,142 & 437,513 & 24,426  \\ \midrule
        \multirow{2}{*}{Det./Seg.}   &LVIS v0.5~\cite{gupta2019lvis} 	& 1,230 & 57,000 &20,000  \\ 
         &LVIS v1~\cite{gupta2019lvis} 	& 1,203 &  100,000 &19,800 	\\  \midrule
        \multirow{2}{*}{Multi-label Cls.}   &VOC-LT~\cite{wu2020distribution} 	& 20 & 1,142 &4,952 	 \\ 
         &COCO-LT~\cite{wu2020distribution} 	& 80 &  1,909 &5,000	 \\ \midrule
         Video Cls.   &VideoLT~\cite{zhang2021videolt} 	& 1,004 & 179,352 & 51,244  	 \\ 
        \bottomrule
	\end{tabular}  
    \end{threeparttable}}
    \end{center}   \vspace{-0.15in}
\end{table}

\subsection{Datasets}\label{sec_dataset}
In recent years, a variety of visual datasets have been released  for  long-tailed learning, differing in tasks, class numbers and sample numbers. In Table~\ref{dataset}, we summarize nine visual datasets that are  widely used in the deep long-tailed learning community. 

In long-tailed  image classification, there are four benchmark  datasets: ImageNet-LT~\cite{liu2019large}, CIFAR100-LT~\cite{cao2019learning}, Places-LT~\cite{liu2019large}, and iNaturalist 2018~\cite{van2018inaturalist}. The previous three  are sampled  from ImageNet~\cite{deng2009imagenet}, CIFAR100~\cite{krizhevsky2009learning} and Places365~\cite{zhou2014learning} following Pareto distributions, respectively, while  iNaturalist   is a real-world long-tailed dataset. The imbalance ratio of ImageNet-LT, Places-LT and iNaturalist   are  256, 996 and 500, respectively;  CIFAR100-LT has three variants with   various imbalance ratios $\{10,50,100\}$. 

In long-tailed object detection and instance segmentation, LVIS~\cite{gupta2019lvis}, providing precise bounding box and mask annotations, is the widely-used benchmark.  In multi-label image classification, the benchmarks are VOC-LT~\cite{wu2020distribution} and COCO-LT~\cite{wu2020distribution}, which are sampled from PASCAL VOC 2012~\cite{everingham2015pascal} and COCO~\cite{lin2014microsoft}, respectively. Recently, a   large-scale ``untrimmed" video dataset, namely VideoLT~\cite{zhang2021videolt}, was released for long-tailed video recognition.


\subsection{Evaluation Metrics}
 
{Long-tailed learning seeks to train a well-performing model on the data with long-tailed class imbalance. To evaluate how well    class imbalance is resolved,   the  model performance on all classes  and the performance  on      class subsets (\ie head, middle and tail classes) are usually reported. Note that  the evaluation metrics should treat each class equally. Following this principle,  top-1 accuracy  or error rate   is often used  for  balanced test sets, where every test sample is equally important. When the test set is not balanced,  mean Average Precision (mAP)   or macro accuracy       is often adopted since the two metrics   treat  each class   equally. For example, in previous studies, top-1 accuracy  or error rate  was  widely used  for long-tailed image classification, in which the test set is usually assumed to be near-balanced. Meanwhile, mAP  was adopted for long-tailed object detection, instance segmentation and multi-label image classification, where the test set is usually not balanced.}  

\subsection{{Applications}}\label{Sec5}
 {The main   applications of deep long-tailed learning  include image classification,   detection  segmentation, and visual relation learning.}
 
{\textbf{Image Classification}.
The most common applications of long-tailed learning are multi-class classification~\cite{liu2019large,kang2019decoupling,zhou2020bbn,tang2020long} and multi-label classification~\cite{wu2020distribution,guo2021long}. As mentioned in Section~\ref{sec_dataset}, there are many artificially sampled long-tailed datasets from  widely-used  multi-class classification datasets (\ie ImageNet, CIFAR, and Places) and multi-label classification datasets (\ie VOC and COCO). Based on these datasets, various long-tailed learning methods have been proposed, as shown in Section~\ref{sec_method}. Besides these artificial tasks, long-tailed learning  is also applied to  real-world applications, including species classification~\cite{van2018inaturalist,miao2021iterative,keaton2021fine}, face recognition~\cite{zhang2017range,cao2020domain,zhong2019unequal,liu2020deep}, face attribute classification~\cite{dong2017class},   cloth attribute classification~\cite{dong2017class}, age classification~\cite{deng2021pml},   rail surface defect  detection~\cite{zhang2021rail}, and medical image diagnosis~\cite{ju2021relational,galdran2021balancedmixup}.   These real applications usually require   more fine-grained discrimination abilities, since the differences among their classes are more subtle. Due to this  new challenge,  existing deep long-tailed learning methods tend to fail in these  applications, since they   only focus on addressing the  class imbalance and cannot essentially identify subtle class differences. Therefore, when  exploring new methods to   handle these applications, it is worth considering how to tackle the challenges of  class imbalance and fine-grained information identification, simultaneously.}
  
{\textbf{Image Detection / Segmentation}.
Object detection and instance segmentation has attracted increasing attention in the long-tailed learning community~\cite{lin2017focal,hsieh2021droploss,li2020overcoming,weyand2020google,zang2021fasa,wu2020forest}, where most existing studies are conducted based on LVIS and COCO. In addition to these widely-used benchmarks, many other applications  have also been explored, including urban scene understanding~\cite{he2021re,mao2021one} and unmanned aerial vehicle detection~\cite{yu2021towards}. Compared to artificial tasks on LVIS and COCO, these real applications are more challenging due to more complex environments in the wild. For example, the images may be collected from different weather conditions  or different times in a day, which may  lead to multiple image domains with different data distributions and inconsistent class skewness. When facing these new challenges, existing deep long-tailed learning methods tend to fail. Hence, it is worth exploring how to simultaneously resolve the challenges of    class imbalance and domain shifts for handling these applications.} 
   
{\textbf{Visual Relation Learning}.
Visual relation learning is important   for image understanding and is attracting rising attention in the long-tailed learning community. Important applications include long-tailed scene graph generation~\cite{Desai2021Learning,dhingra2021bgt}, long-tailed visual question answering and image captioning~\cite{chen2021reltransformer,licalibrating}.  Most existing  long-tailed studies focus on discriminative tasks, so they cannot be    applied to the aforementioned applications that require modeling relations between objects or those between images and texts. Even so, it is interesting to explore the  high-level ideas (\eg class re-balancing) in existing long-tailed studies to design application-customized  approaches for visual relation learning.}
 



\subsection{Relationships with Related Tasks}
{We then briefly discuss several related tasks, including non-deep long-tailed learning, class-imbalanced learning, few-shot learning,   and out-of-domain generalization.} 

{\textbf{Non-deep long-tailed learning}. There are  a lot of non-deep learning approaches for long-tailed problems~\cite{wang2010comparative,loy2012stream,yang2014context}. They usually explore   prior knowledge to enhance classic machine learning algorithms for handling the long-tailed problem. For example, the prior of similarity   among categories  is used to regularize kernel machine algorithm for long-tailed object recognition~\cite{wang2010comparative}. Moreover, the prior of a long-tailed power-law distribution   produced by  the Pitman-Yor Processes (PYP) method~\cite{pitman1997two} is applied to enhance  the Bayesian non-parametric framework for long-tailed active learning~\cite{loy2012stream}.  An artificial  distribution prior is adopted to construct tail-class data augmentation to enhance KNN and SVM for long-tailed scene parsing~\cite{yang2014context}. Almost all these approaches extract image features based on Scale Invariant Feature Transform (SIFT)~\cite{lowe2004distinctive},  Histogram of Gradient Orientation (HOG)~\cite{dalal2005histograms}, or RGB color histogram~\cite{swain1991color}. 
Such   representation approaches, however, cannot extract highly informative and discriminative features for  real  visual applications~\cite{lecun2015deep} and thus   lead to limited performance in long-tailed learning. Recently, in light of the powerful abilities of deep   networks for image representation,  deep long-tailed methods have achieved significant performance improvement for long-tailed learning. More encouragingly, the use of deep   networks also inspires plenty of new solution paradigms for long-tailed learning, such as transfer learning, decoupled training and ensemble learning, which will be introduced in the next section.}  

{\textbf{Class-imbalanced learning}~\cite{he2009learning,wang2020deep} also seeks to train models from class-imbalanced samples. In this sense, long-tailed learning can be regarded as a   challenging sub-task of  class-imbalanced learning. The dominant distinction is that the classes of  long-tailed learning follow a \emph{long-tailed class  distribution}, which is not necessary for   class-imbalanced learning. More differences include that  in long-tailed learning the number of classes is usually large and the tail-class samples are often very scarce, whereas the number of minority-class samples  in class-imbalanced learning is not necessarily small in an absolute sense.  These extra challenges lead long-tailed learning to  be   a more challenging  task than class-imbalanced learning. Despite these differences,  both    seek to resolve the class imbalance, so some high-level solution ideas  (\eg class re-balancing) are  shared between them.}

\textbf{Few-shot learning}~\cite{snell2017prototypical,sung2018learning,sun2019meta,wang2020generalizing} aims to  train models from a limited number of labeled samples (\eg  1 or 5) per class. In this regard, few-shot learning can be regarded as a sub-task of long-tailed learning, in which the tail classes generally have a very  small number of samples.

\textbf{Out-of-domain Generalization}~\cite{krueger2021out,shen2021towards} indicates a class of tasks, in which the  training distribution is inconsistent  with the unknown test distribution. Such inconsistency includes inconsistent data marginal distributions (e.g., domain adaptation~\cite{pan2010domain,tzeng2017adversarial,zhang2019whole,zhang2020collaborative,qiu2021source,wu2021heterogeneous} and domain generalization~\cite{li2017deeper,li2018domain}), inconsistent class distributions (e.g., long-tailed learning~\cite{liu2019large,kang2019decoupling,jamal2020rethinking}, open-set learning~\cite{neal2018open,fu2019vocabulary}), and the combination of the previous two situations. From this perspective, long-tailed learning can be viewed as  a specific task within out-of-domain generalization.

\begin{table*}[htp]   
	{\caption{Summary of existing  deep long-tailed learning methods  published in the top-tier conferences  before mid-2021. There are three main categories: class re-balancing, information augmentation and module improvement. In this table, ``CSL" indicates class-sensitive learning;  ``LA" indicates logit adjustment;  ``TL" represents transfer learning; ``Aug" indicates data augmentation;  ``RL" indicates representation learning; ``CD" indicates classifier design, which seeks to design new classifiers or prediction schemes for long-tailed recognition; ``DT" indicates  decoupled training, where the feature extractor and the classifier are trained separately;  ``Ensemble" indicates ensemble learning based methods. In addition, ``Target Aspect" indicates  from which  aspect  an approach seeks to resolve the class imbalance. We also make   our codebase and our collected long-tailed learning resources  available at \url{https://github.com/Vanint/Awesome-LongTailed-Learning}.}\label{Taxonomy}} 
 \vspace{-0.1in}
 \begin{center}
 \begin{threeparttable} 
    \resizebox{0.99\textwidth}{!}{
 	\begin{tabular}{lccccccccccccc}\toprule  
        \multirow{2}{*}{Method}   &      \multirow{2}{*}{Year} &   \multicolumn{3}{c}{Class Re-balancing}  &  &   \multicolumn{2}{c}{Augmentation} & &    \multicolumn{4}{c}{Module  Improvement} &   \multirow{2}{*}{Target Aspect}  \cr  \cmidrule{3-5}    \cmidrule{7-8}  \cmidrule{10-13}  
        & &   Re-sampling  & CSL  & LA && TL & Aug && RL &CD   & DT  &Ensemble  \cr
        \midrule
      LMLE~\cite{huang2016learning} &  2016 &  &   &    && &  &&  \cmark  &    & & & feature \\ 
      HFL~\cite{ouyang2016factors}  &   2016 &  &    & &&    &  && \cmark & & & & feature  \\ 
      Focal loss~\cite{lin2017focal} &  2017 &    & \cmark   &  &&    & &&  &   &  &  & objective   \\ 
      Range loss~\cite{zhang2017range} &  2017 &    &   & &&    & &&  \cmark &  & & & feature  \\ 
      CRL~\cite{dong2017class} &   2017 &    &    & &&    & && \cmark  & & & & feature  \\
      MetaModelNet~\cite{wang2017learning}  &   2017 &   &    &   &&  \cmark &  && &  &   & &    \\ 
      DSTL~\cite{cui2018large} &   2018 &   &    &   &&  \cmark & &&  & &  &   \\ 
      DCL~\cite{wang2019dynamic} & 2019 & \cmark   &     &   &&   &  & & &    & &  & sample \\ 
      
      Meta-Weight-Net~\cite{shu2019meta}&  2019 &   &\cmark     &   &&   &  & &  &   &   & &  objective  \\ 
      LDAM~\cite{cao2019learning} & 2019 &   &  \cmark   &   &&   &  & &   &  &  &  &  objective  \\ 
      CB~\cite{cui2019class}& 2019 &   & \cmark &   && & &&  & &   &   & objective  \\

      UML~\cite{khan2019striking}    & 2019 &   & \cmark    &   && &  &&  & &   &  & feature  \\ 
      FTL~\cite{yin2019feature}    & 2019 &   &    &   &&  \cmark &\cmark  && &   & & & feature  \\ 

      Unequal-training~\cite{zhong2019unequal}   & 2019 &   &     &   &&  &  &&\cmark & &  & & feature  \\ 
      
      OLTR~\cite{liu2019large}    & 2019 &   &    &   &&  &  && \cmark  & &  & & feature  \\

      Balanced Meta-Softmax~\cite{jiawei2020balanced}  &  2020 &  \cmark   &  \cmark       & &&   &   &&    &   &   &   & sample, objective      \\
      Decoupling~\cite{kang2019decoupling}& 2020 &  \cmark &\cmark     &   &&   &  && \cmark  &  
      \cmark  &  \cmark  &   & feature, classifier   \\ 
      
      LST~\cite{hu2020learning} &   2020 & \cmark  &      &   &&  \cmark  &   &&  &    &   &   & sample   \\

      Domain adaptation~\cite{jamal2020rethinking}&  2020 &    &\cmark     &   &&  &  && &  &      &   &  objective\\ 
      Equalization loss (ESQL)~\cite{tan2020equalization}  & 2020 &   &  \cmark     &   &&   &  &&  &  &        & &  objective \\ 
      DBM~\cite{cao2020domain}&   2020 &    & \cmark     &   &&   &  && &  &   &   &  objective \\ 
      
      Distribution-balanced loss~\cite{wu2020distribution} &    2020 &   &\cmark       &   &&  &    &&    &   &   &     &  objective    \\ 
      
       UNO-IC~\cite{tian2020posterior}  & 2020 &   &         &  \cmark   &&   &    &&    & &   &  & prediction \\  
      
      De-confound-TDE~\cite{tang2020long}  &   2020 &   &      &  \cmark  &&     &    &&   &  \cmark   & & &  prediction \\ 
      
      M2m~\cite{kim2020m2m} & 2020 &   &      &   &&    \cmark& \cmark &&  &  &    &    & sample  \\ 
      LEAP~\cite{liu2020deep} &  2020 &   &      &   &&  \cmark  & \cmark &&  \cmark  &  & &   & feature   \\ 
      OFA~\cite{chu2020feature} &  2020 &   &      &   &&  \cmark  & \cmark  && &   & \cmark   &      & feature   \\ 
      
      SSP~\cite{yang2020rethinking}  &   2020 &   &      &   &&   \cmark  &     && \cmark   & &   &   & feature      \\ 
    
      LFME~\cite{xiang2020learning} &    2020 &    &      &   &&  \cmark  &   &&  &    &     &   \cmark & sample, model \\ 
      IEM~\cite{zhu2020inflated} &   2020 &   &      &   &&    &   &&  \cmark &  &    &  & feature  \\  

      Deep-RTC~\cite{wu2020solving} &   2020 &   &      &   &&     &    &&   &  \cmark     & &    & classifier  \\ 
        SimCal~\cite{wang2020devil} & 2020 &    &      &   &&     &    &&  &    &    \cmark  & \cmark   & sample, model \\  
      BBN~\cite{zhou2020bbn}&  2020 &    &     &   &&   &  &&  &  &  &  \cmark & sample,  model \\ 
      BAGS~\cite{li2020overcoming}  & 2020 &      &      &   &&    &   &&   &  &      &  \cmark & sample, model \\ 
      
      VideoLT~\cite{zhang2021videolt}&   2021 &   \cmark   &       &   && &       &&   &  &   & & sample \\ 
      LOCE~\cite{feng2021exploring}&  2021 &   \cmark   & \cmark      &   && &       &&   &  &   & & sample, objective \\  
      DARS~\cite{he2021re} &  2021 &  \cmark &    \cmark   &   && \cmark  &     &&   &  &  &  & sample, objective \\
      CReST~\cite{wei2021crest}  &   2021 &  \cmark   &     &   &&    \cmark  &   &&   &   & & & sample\\      
      GIST~\cite{liu2021gistnet}&  2021 &  \cmark  &       &   &&  \cmark &      &&  & \cmark   & & & classifier  \\
      FASA~\cite{zang2021fasa}&  2021 &  \cmark  &       &   &&  &     \cmark  & &    &  & & & feature  \\

      Equalization loss v2~\cite{tan2021equalization} &  2021 &   &  \cmark     &   &&   &  &&  &  &        &  &  objective \\     
      Seesaw loss~\cite{wang2021seesaw} &   2021 &   &  \cmark     &   &&   &  &&  &  &        &  &  objective \\   
      ACSL~\cite{wang2021adaptive} &   2021 &   &  \cmark     &   &&   &  &&  &  &        & &  objective\\   
      
      IB~\cite{Influence2021Park} &  2021 &   &    \cmark    &   &&  &     &&   &  &  & &  objective \\
      PML~\cite{deng2021pml}  &   2021 &   &  \cmark      &   &&   &   &&  & &     & &  objective \\ 
      VS~\cite{kini2021label}&   2021 &   &    \cmark   &    & & &      &&  &  &   &   &   objective   \\
      LADE~\cite{hong2020disentangling}  &  2021 &   & \cmark       &   \cmark  &&   & &&     &   &     &  &  objective, prediction  \\
 
      RoBal~\cite{wu2021adversarial}  &  2021 &   &   \cmark     &   \cmark  &&   &   &&  & \cmark &   &   & objective, prediction   \\  
      DisAlign~\cite{zhang2021distribution}  &   2021 &   & \cmark       &   \cmark  && &  &&     &    &  \cmark  &   & objective, classifier   \\       
      MiSLAS~\cite{zhong2021improving}  &  2021 &   & \cmark       &   &&   &    \cmark   &&   &   & \cmark  &  & objective, feature, classifier \\

      Logit adjustment~\cite{menon2020long}  &  2021 &   &        &   \cmark &&  &    && &   &   &     &   prediction   \\

      Conceptual 12M~\cite{changpinyo2021conceptual}  &   2021 &   &     &   &&    \cmark  &   &&   &   & & \\ 
    
      DiVE~\cite{he2021distilling}&   2021 &   &       &   &&  \cmark &      &&  &  & &   \\
      MosaicOS~\cite{zhang2021mosaicos} &  2021 &   &       &   &&  \cmark &      &&  &  & &   \\

      RSG~\cite{wang2021rsg}  &  2021 &   &       &   &&  \cmark  &    \cmark   &&   &  & & & feature \\
      SSD~\cite{li2021self}&  2021 &   &       &   &&  \cmark &      &&  &  &  \cmark  &   \\
      RIDE~\cite{wang2020long}  &   2021 &   &       &   &&   \cmark &     &&    &  &   &   \cmark & model  \\ 
      MetaSAug~\cite{li2021metasaug}  &  2021 &   &       &   &&   &    \cmark   &&   &  & & & sample \\
      PaCo~\cite{cui2021parametric}&  2021 &   &       &   && &      && \cmark   &  &   & & feature  \\
      DRO-LT~\cite{samuel2021distributional}&  2021 &   &       &   && &      && \cmark   &  &   & & feature \\
      Unsupervised discovery~\cite{weng2021unsupervised}  & 2021 &   &       &   &&   &      && \cmark  &  &  & & feature  \\
      Hybrid~\cite{wang2021contrastive}  &  2021 &   &       &   &&   &   &&  \cmark  & &    & & feature  \\   

      KCL~\cite{kang2021exploring}  &   2021 &   &      &   &&      &     && \cmark   &  & \cmark&    & feature  \\ 
      DT2~\cite{Desai2021Learning}&   2021 &   &       &   && &      &&  &  & \cmark  &  &   feature, classifier   \\
      LTML~\cite{guo2021long}  &  2021 &      &     &     &&   &   &&    &   & & \cmark & sample, model \\ 
      ACE~\cite{cai2021ace}&  2021 &      &       &   && &       &&  &  &   & \cmark & sample, model \\      
      ResLT~\cite{cui2021reslt}&   2021 &    &       &    & & &      &&  &  &   & \cmark & sample, model \\
      SADE~\cite{zhang2021test}&   2021 &   &      &     & & &      &&  &  &   &  \cmark & objective, model  \\
     \bottomrule
	\end{tabular}}
	 \end{threeparttable}
	 \end{center} 
\end{table*} 

\section{Classic Methods}\label{sec_method}
As shown in Fig.~\ref{fig_taxonomy}, we divide existing  deep long-tailed learning methods    into three main categories {according to their main technical characteristics}, including class re-balancing, information augmentation, and module improvement.
More specifically, class re-balancing  consists of three sub-categories: re-sampling, class-sensitive  learning (CSL), and logit adjustment (LA). Information augmentation comprises transfer learning (TL) and data augmentation (Aug). Module improvement includes representation learning (RL), classifier design (CD), decoupled training (DT) and ensemble learning (Ensemble). According to this taxonomy, we sort out existing   methods in Table~\ref{Taxonomy} and   review them in detail as follows.

\subsection{{Class Re-balancing}}
{Class re-balancing, a mainstream paradigm  in long-tailed learning, seeks to re-balance the negative influence brought by the class  imbalance in training sample numbers. This type of methods has three main sub-categories: re-sampling, class-sensitive  learning, and logit adjustment.} We begin with re-sampling based methods, followed by class-sensitive learning and logit adjustment. 
 
\subsubsection{Re-sampling}\label{sec_resampling}

{Conventional training of deep networks is based on mini-batch gradient descent with random sampling, \ie each sample has an equal probability of being sampled. Such a sampling manner, however,  ignores the  imbalance issue in   long-tailed learning,  and  naturally  samples more head-class samples than tail-class samples in each sample mini-batch. This makes the resulting deep models  biased  towards head  classes and perform  poorly on tail classes. To address this issue, re-sampling~\cite{chawla2002smote,estabrooks2004multiple,liu2008exploratory,zhang2021learning} has been explored   to re-balance classes by adjusting the  number of samples per class  in each sample  batch for model training.}

 {In the non-deep learning era, the most classic re-sampling approaches are random over-sampling (ROS) and random under-sampling (RUS). Specifically,  ROS randomly repeats the samples from   minority classes   to re-balance  classes   before   training, while RUS  randomly discards the samples from  majority  classes. When applying them to deep long-tailed learning where the classes are highly skewed,  ROS with duplicated tail-class data might lead to overfitting over   tail   classes, while RUS  might  discard precious head-class samples and    degrade   model performance on head classes~\cite{zhou2020bbn}.  Instead of using random re-sampling, recent deep long-tailed  studies have developed various class-balanced  sampling methods for mini-batch  training of deep models.}

{We begin with Decoupling~\cite{kang2019decoupling},  in which four   sampling strategies were evaluated  for representation learning of long-tailed data, including  random sampling, class-balanced sampling, square-root sampling and progressively-balanced sampling. Specifically, class-balanced sampling means that each class has an equal probability of being selected. Square-root sampling~\cite{mahajan2018exploring} is a variant of class-balanced sampling, where the sampling probability  of each class is related to the square root of the sample size in the corresponding class. Progressively-balanced sampling~\cite{kang2019decoupling}  interpolates progressively between  random  and class-balanced sampling. Based on empirical results, Decoupling~\cite{kang2019decoupling} found that square-root sampling and progressively-balanced sampling are better strategies for standard model training   in long-tailed recognition.   The two strategies, however, require knowing the training sample frequencies of different classes in advance, which may be unavailable in real applications.}

{To address the above issue, recent studies proposed various adaptive  sampling strategies. Dynamic Curriculum Learning (DCL)~\cite{wang2019dynamic}  developed a new curriculum strategy to dynamically sample data for class re-balancing. The basic idea is that  the more instances from one class are sampled as training proceeds, the  lower  probability of   this class would be sampled in later stages. Following this idea, DCL   first conducts random sampling to learn general representations, and then samples more tail-class instances based on the     curriculum strategy to handle the   imbalance.   In addition to using the accumulated sampling times,  Long-tailed Object Detector with Classification Equilibrium (LOCE)~\cite{feng2021exploring}   proposed to  monitor  model training on different classes via the \emph{mean   classification prediction score} (\ie running prediction probability), and used this score to guide the sampling rates for different classes. Furthermore, VideoLT~\cite{zhang2021videolt}, focusing on long-tailed video recognition,   introduced  a new FrameStack   method that dynamically adjusts the sampling rates of different classes based on   \emph{running model performance} during training, so that it can sample more video frames from tail classes (generally with lower running performance).} 

{Besides using the statistics computed during model training, some re-sampling approaches    resorted to meta learning~\cite{hospedales2021meta}.
Balanced Meta-softmax~\cite{jiawei2020balanced} developed a  meta-learning-based  sampling method to estimate the optimal sampling rates of different classes for long-tailed learning.  Specifically, the developed meta learning method seeks to learn the best sample distribution parameter by optimizing the \emph{model classification  performance} on a balanced \emph{meta} validation set. Similarly, Feature Augmentation and Sampling Adaptation (FASA)~\cite{zang2021fasa} explored the \emph{model classification loss} on a   balanced \emph{meta} validation set as a score, which is used to adjust the sampling rate for different classes so that the under-represented tail classes can be sampled more.}

{Note that  some long-tailed visual tasks may have multiple levels of imbalance. For example, long-tailed instance segmentation is imbalanced in terms of both   images and   instances (\ie the number of instances per image is also imbalanced). To address this task, Simple Calibration (SimCal)~\cite{wang2020devil}   proposed a new bi-level class-balanced sampling strategy that combines image-level  and instance-level  re-sampling for class re-balancing. }

{\textbf{Discussions}. Re-sampling methods seek to address the class imbalance issue at the sample level. When the label frequencies of different classes are known a priori,   progressively-balanced sampling~\cite{kang2019decoupling} is recommended. Otherwise, using the statistics   of model training to guide   re-sampling~\cite{feng2021exploring} is a preferred solution for real applications. For meta-learning-based re-sampling, it may be difficult to construct a meta validation set in real scenarios. Note that if   one re-sampling strategy has already  addressed class imbalance well,  further using other re-sampling methods may not bring extra benefits. Moreover, the high-level ideas of these re-sampling methods can be applied to design multi-level re-sampling strategies if there 
are multiple levels of imbalance in real applications.}

\subsubsection{{Class-sensitive Learning}}
{Conventional training of deep networks is based on the softmax cross-entropy  loss (c.f. Table~\ref{losses}). This loss  ignores the class imbalance in data sizes and   tends to generate uneven gradients for different classes. 
That is, each positive sample of one class can be seen as a negative sample for other classes in cross-entropy, which leads head classes to receive more supporting gradients (as they usually are positive samples) and causes tail classes to receive more suppressed gradients (as they usually are negative samples) ~\cite{tan2020equalization,hsieh2021droploss}.  
To address this, class-sensitive learning seeks to particularly adjust the training loss values for various classes  to re-balance the uneven training  effects   caused by the imbalance issue~\cite{elkan2001foundations,zhou2005training,zhao2018adaptive,zhang2018online,zhang2019online,sun2007cost}. There are  two main types of class-sensitive strategies,   \ie   re-weighting and  re-margining.  We begin with class re-weighting as follows.}

\begin{table}[t] 
	\caption{Summary of   losses. In this table, $z$ and $p$ indicate the predicted logits  and the softmax probability of the sample $x$, where $z_y$ and $p_y$ correspond to the class $y$. Moreover, $n$ indicates the total number of training data,  where $n_y$ is the sample number of the class $y$. In addition, $\pi$ denotes the vector of sample frequencies, where  $\pi_y\small{=}n_y/n$ represents the label frequency of the class $y$. The  class-wise weight is denoted by $\omega$ and the class-wise margin is denoted by $\Delta$, if no more specific value is given. Loss-related  parameters include $\gamma$.} 
	\label{losses} 
	\vspace{-0.15in}
 \begin{center}
 \begin{threeparttable} 
    \resizebox{0.48\textwidth}{!}{
 	\begin{tabular}{llc}\toprule  
         
     Loss    &  Formulation & Type \cr
      
     \midrule
      Softmax loss & ${\mathcal{L}_{\rm ce}} =  - \log (p_y)$ & -  \\ 
      Focal loss~\cite{lin2017focal}  & ${\mathcal{L}_{\rm fl}} =  - (1-p_y)^{\gamma} \log (p_y)$   & re-weighting  \\  
      Weighted Softmax loss & ${\mathcal{L}_{\rm wce}} =  -\frac{1}{\pi_y} \log (p_y)$ & re-weighting \\ 
      Class-balanced loss~\cite{cui2019class}& ${\mathcal{L}_{\rm cb}} =  -  \frac{1-\gamma}{1-\gamma^{n_y}} \log (p_y)$ & re-weighting\\  
      Balanced Softmax loss~\cite{jiawei2020balanced} & ${\mathcal{L}_{\rm bs}} =  - \log (\frac{\pi_y \exp(z_y)}{\sum_j \pi_j \exp(z_j) })$  &  re-weighting\\
      Equalization loss~\cite{tan2020equalization} & ${\mathcal{L}_{\rm eq}} =  - \log (\frac{ \exp(z_y)}{\sum_j \omega_j \exp(z_j) })$   &  re-weighting   \\  
      LDAM  loss~\cite{cao2019learning} & ${\mathcal{L}_{\rm ldam}} =  - \log (\frac{ \exp(z_y -\Delta_y)}{\sum_j  \exp(z_j-\Delta_j) })$ &  re-margining   \\   
     \bottomrule
	\end{tabular}}
	 \end{threeparttable}
	 \end{center} 
	\vspace{-0.1in}
\end{table}  

{\textbf{Re-weighting.} To address the class imbalance, re-weighting attempts to adjust the training loss values for different classes by multiplying them with  different weights.  The most intuitive  method is to directly use the \emph{label frequencies of training samples} for loss   re-weighting to re-balance the uneven positive gradients among classes. For example,  weighted softmax   (c.f. Table~\ref{losses})  directly multiplies the loss values of different classes by the   inverse of training label frequencies. However, simply multiplying by its inverse may not be the  optimal solution. Recent studies thus proposed to tune the  influence of training label frequencies  based on sample-aware influences~\cite{Influence2021Park}. Moreover, Class-balanced loss (CB)~\cite{cui2019class} introduced a novel concept of \emph{effective number} to approximate the expected sample number of different classes, which is an exponential function of their  training label number. Following this, CB loss enforces a class-balanced re-weighting term, inversely proportional to the effective number of classes, to address the class imbalance (c.f. Table~\ref{losses}).}
{Besides the aforementioned re-weighting at the level  of   log probabilities, we can also use the training label frequencies to re-weight prediction logits.  Balanced Softmax~\cite{jiawei2020balanced}  proposed to   adjust prediction logits  by multiplying by    the  label  frequencies, so that the bias of   class   imbalance can be  alleviated by the label prior before computing final losses. Afterwards, Vector-scaling loss  (VS)~\cite{kini2021label}  intuitively analyzed the distinct effects of additive and multiplicative logit-adjusted losses, leading to a novel VS loss to combine the advantages of both forms of adjustment.}

{Instead of using training label frequencies,  Focal loss~\cite{lin2017focal} explored \emph{class prediction hardness}   for re-weighting. This is inspired by the observation that \emph{class imbalance usually increases the prediction hardness of tail classes, whose prediction probabilities would be lower than those of head classes}. Following this, Focal loss uses the prediction probabilities to  inversely re-weight classes (c.f. Table~\ref{losses}), so that it can assign higher weights to the harder tail classes but lower weights to the easier head classes.    
Besides using a pre-defined weighting function, the class  weights can  also  be   learned from data. For instance, Meta-Weight-Net~\cite{shu2019meta} proposed to learn an MLP-approximated weighting function based on a balanced validation  set for class-sensitive learning.}



{Some recent studies~\cite{wu2020distribution,tan2020equalization} also seek to address the negative gradient over-suppression issue of tail classes. For example,  Equalization loss~\cite{tan2020equalization}   directly down-weights the loss values of tail-class samples when they serve as negative labels for   head-class samples. However, simply down-weighting negative gradients may harm the discriminative abilities of deep models. To address this, Adaptive Class Suppression loss (ACSL)~\cite{wang2021adaptive}    uses the \emph{output confidence} to decide whether to suppress the gradient for a negative label. Specifically, if the prediction probability of a negative label is larger than a pre-defined threshold,  it means that the model is confused about this class so the weight for this class is set to 1 to improve model discrimination; otherwise, the weight is set to 0 to avoid   negative over-suppression.     Moreover, Equalization loss v2~\cite{tan2021equalization}   extended the equalization loss~\cite{tan2020equalization}  by   introducing a novel gradient-guided re-weighting mechanism that dynamically  up-weights the positive gradients and down-weights the negative gradients for different classes.
Similarly, Seesaw loss~\cite{wang2021seesaw} re-balances positive and negative gradients for each class with two  re-weighting factors, \ie mitigation and compensation. Specifically, to address gradient over-suppression, the mitigation factor alleviates the penalty to tail classes  based on a   dynamically  cumulative sampling number of different classes. Meanwhile, if a false positive sample  is observed, the compensation factor up-weights the penalty to the corresponding class for improving  model discrimination.}

{\textbf{Re-margining}. To  handle the class imbalance, re-margining attempts to  adjust losses by subtracting different margin factors for different classes, so that they have a different  minimal margin (\ie distance)  between features and the  classifier. Directly using existing soft margin losses~\cite{wang2018additive,koltchinskii2002empirical} is unfeasible, since they ignore the issue of class imbalance. To address this, Label-Distribution-Aware Margin (LDAM)~\cite{cao2019learning}  enforces class-dependent margin factors for different classes based on  their training label frequencies, which encourages     tail classes to have larger margins.}
 
{However, the training label frequencies may be unknown  in real applications,  and simply using them for re-margining also ignores the  status of model training  on different classes. To address this, recent studies   explored various adaptive re-margining methods. Uncertainty-based margin learning (UML)~\cite{khan2019striking}  found that \emph{the class prediction uncertainty is inversely proportional to the  training label   frequencies, i.e., tail classes are more uncertain}. Inspired by this, UML proposed to use the estimated class-level uncertainty to re-margin losses, so that the tail classes with higher class uncertainty    incur a higher loss value and thus have a  larger margin between   features and the classifier. 
Moreover, LOCE~\cite{feng2021exploring}  proposed to use the \emph{mean class prediction score} to monitor the learning status of different classes and apply it to guide class-level margin adjustment for enhancing tail classes. Domain balancing~\cite{cao2020domain}   introduced a novel  frequency indicator based on the \emph{inter-class compactness of features}, and uses this indicator to re-margin the feature space of tail domains.} 
{Despite effectiveness,   the above re-margining methods for  encouraging large tail-class margins may degrade the feature learning of head classes. To address this, RoBal~\cite{wu2021adversarial}   further enforces a  margin factor  to also enlarge head-class margins.}

{\textbf{Discussions}. These class-sensitive learning methods aim to resolve the class imbalance issue at the objective level. We summarize some of them in Table~\ref{losses}. Both re-weighting and re-margining methods have a similar effect on re-balancing   classes.  If the negative influence of class imbalance can be  addressed  by one class-sensitive  approach well, it is unnecessary to further apply other class-sensitive methods, which would not bring further performance gain and even harm   performance. More specifically, if the training label frequencies are available, directly using them for re-weighting (\eg Balanced Softmax~\cite{jiawei2020balanced} and VS~\cite{kini2021label}) or re-margining (\eg LDAM~\cite{cao2019learning}) provides a simple and generally effective solution for real applications. If not, it is preferred to use  the mean class prediction score  to guide class-sensitive learning (\eg ACSL~\cite{wang2021adaptive} and LOCE~\cite{feng2021exploring}) thanks to its simplicity. One can also consider other guidance, like intra-class compactness. However, inter-class compactness of features~\cite{cao2020domain} may be not that informative when the feature dimensions are very high, while the prediction uncertainty~\cite{khan2019striking} may be difficult to  estimate accurately in practice. Moreover, using prediction hardness for re-weighting in Focal loss  performs well when the number of classes is not large,  but may fail when facing a large number of classes. Furthermore,  Equalization loss v2, Seesaw loss  and RoBal can also be considered if   the challenges that they try to resolve appear in real applications.}

\subsubsection{Logit Adjustment}  
{Logit adjustment~\cite{provost2000machine,menon2020long} seeks to resolve the class  imbalance by   adjusting the prediction logits of a  class-biased  deep model. 
One recent study~\cite{menon2020long} comprehensively analyzed  logit adjustment  via   training label frequencies of different classes  in long-tailed recognition, and theoretically showed that \emph{logit adjustment is Fisher consistent to minimize the average per-class error}. 
Following this idea, RoBal~\cite{wu2021adversarial} applied a post-processing strategy to  adjust the cosine classifier  based on  training label frequencies.}

{However, the above methods tent to fail when the training label frequencies are unavailable. To address this this, UNO-IC~\cite{tian2020posterior} proposed to learn the logit offset based on a \emph{balanced} meta validation set and use it to calibrate  the biased model predictions.   Instead of using a meta validation set, DisAlign~\cite{zhang2021distribution} applied an adaptive calibration function for logit adjustment, where the  calibration function is learned by matching the calibrated prediction distribution to a pre-defined relatively balanced class distribution.  
}

{The idea of logit adjustment naturally suits   agnostic test class distributions. If the test label frequencies are available, LADE~\cite{hong2020disentangling} proposed to use them to post-adjust model outputs so that the trained model can be calibrated for arbitrary  test class distributions. However, the test label frequencies are usually unavailable, which makes LADE less practical in real scenarios.}

{\textbf{Discussions}. To summarize, these logit adjustment methods address the class imbalance  at the prediction level. If the training label  frequencies  are known, directly using them to post-adjust the predictions of   biased deep models  is recommended~\cite{menon2020long,wu2021adversarial}. If such information is unknown, it is preferred to exploit the idea of DisAlign~\cite{zhang2021distribution} to learn an adaptive calibration function.  These logit adjustment methods are  exclusive to each other, so using a well-performing one is  enough for real applications.}

\subsubsection{Summary}\label{sub_class_rebalancing}
{Class re-balancing is relatively simple   among the three main method types of   long-tailed learning,  but  it can    achieve comparable or even better performance. Some methods, especially class-sensitive learning, are theoretically inspired or guaranteed to handle long-tailed problems~\cite{cui2019class,cao2019learning,hong2020disentangling}. These advantages enable  class re-balancing  to be a good candidate for real-world applications.}

{The ultimate goal of its three sub-categories (\ie re-sampling, class-sensitive learning and logit adjustment) are the same, \ie re-balancing classes. Hence, when the class imbalance is not  addressed well, combining them may achieve better performance.   However, these subtypes are sometimes exclusive to each other. For example,  if we have trained a class-balanced deep model via class-sensitive learning, then further using logit adjustment methods to post-adjust model inference will instead lead to biased predictions and suffer poor performance. Therefore, if one wants to combine them, the pipeline should be designed carefully.}

{One drawback of class re-balancing   is that most  methods improve tail-class performance  at the cost of lower  head-class performance, which is like playing on a performance seesaw. Although the overall performance is improved, it cannot essentially handle the issue of lacking  information, particularly  on tail classes due to limited  data sizes. To address this limitation, one feasible solution  is to conduct information augmentation  as follows.}

\subsection{Information Augmentation}\label{sec_IA}
Information augmentation seeks to introduce additional information into model training,  so that the model performance   can be improved for long-tailed  learning. There are two kinds of methods in this method type: transfer learning and data augmentation. 

\subsubsection{Transfer Learning}\label{sec_transfer}
Transfer learning~\cite{pan2009survey,tan2018survey,wang2017learning,wang2021rsg,chu2020feature} seeks to transfer the knowledge from a source domain (\eg datasets) to enhance model training on a target domain. In   long-tailed learning, there are   four main transfer   schemes, \ie  model pre-training,  knowledge distillation, head-to-tail model transfer,  and self-training.

{\textbf{Model pre-training} is a popular  scheme for   deep model training~\cite{erhan2010does,he2019rethinking,hendrycks2019using,zoph2020rethinking,Zhang2021UnleashingTP} and has also been explored in long-tailed learning. For example, Domain-Specific Transfer Learning (DSTL)~\cite{cui2018large} first pre-trains the model with all long-tailed samples for representation learning, and then fine-tunes the   model on a more class-balanced training subset. In this way, DSTL slowly transfers the learned features to tail classes, obtaining more balanced performance among all classes. Rather than supervised pre-training, Self-supervised Pre-training (SSP)~\cite{yang2020rethinking} proposed to first use self-supervised learning (\eg contrastive learning~\cite{he2020momentum} or rotation prediction~\cite{gidaris2018unsupervised})  for model pre-training, followed by standard training   on long-tailed data.  Empirical results show  self-supervised learning helps to learn a balanced feature space for long-tailed learning~\cite{kang2021exploring}. Such a scheme has also been explored to handle long-tailed data with noisy labels~\cite{karthik2021learning}. }

{\textbf{Knowledge distillation} seeks to  train  a student model based on  the outputs of a well-trained teacher model~\cite{hinton2015distilling,gou2021knowledge}. Recent studies have explored  knowledge distillation for long-tailed learning. For example, 
Learning from Multiple Experts (LFME)~\cite{xiang2020learning} first trains multiple experts on several less imbalanced  sample subsets (\eg head, middle and tail sets), and then distills these experts into a  unified student model. Similarly,  Routing Diverse Experts (RIDE)~\cite{wang2020long}   introduced a knowledge distillation method to reduce the parameters of the multi-expert model  by learning a student network with fewer experts. Instead of multi-expert teachers, Distill the Virtual Examples (DiVE)~\cite{he2021distilling} showed  that learning a   class-balanced model  as the teacher   is also beneficial   for long-tailed learning. Following DiVE, Self-Supervision to Distillation (SSD)~\cite{li2021self} developed a new self-distillation scheme to enhance decoupled training (c.f. Section~\ref{sec_decoupled}). Specifically, SSD first trains a calibrated model based on supervised and self-supervised information via the    decoupled training scheme, and then uses the  calibrated model to generate soft labels for all samples. Following that, both the generated soft labels and original long-tailed hard labels are used to distill a new student model, followed by a new classifier fine-tuning stage. }

{\textbf{Head-to-tail model transfer} seeks to transfer the model knowledge from head classes to enhance model performance on tail classes.
For example, MetaModelNet~\cite{wang2017learning}  proposed to learn a meta-network  that can  map few-shot model parameters to many-shot model parameters. To this end,  MetaModelNet first trains a  many-shot model on the head-class training set, and trains  a fake few-shot model on a sampled subset from these classes with a very limited number of data to mimic tail classes. Then, the meta-network is learned by mapping the learned fake few-shot model to the many-shot model. Following that, the learned  meta-network  on head classes is applied to map the true few-shot model trained   on tail classes for obtaining better tail-class performance. Instead of model mapping, Geometric Structure Transfer (GIST)~~\cite{liu2021gistnet} proposed to conduct head-to-tail  transfer  at the classifier level. Specifically, GIST uses  the  relatively large  classifier geometry information   of head classes to   enhance   the tail-class  classifier  weights, so that the performance of tail classes can be improved. }

{\textbf{Self-training} aims to learn well-performing models from a small number of labeled samples and massive unlabeled samples~\cite{zhu2005semi,rosenberg2005semi,wei2021robust}. To be specific, it firstly uses   labeled samples to train  a supervised   model, which is then applied to generate pseudo labels for unlabeled data. Following that, both the labeled  and pseudo-labeled samples are used to re-train  models.  In this way, self-training   can exploit the knowledge from massive unlabeled samples to enhance   long-tailed learning performance.  Such a paradigm, however,    cannot be directly used to handle   long-tailed problems, because both labeled  and   unlabeled datasets may follow long-tailed class distributions with different degrees. In such cases, the trained model on labeled samples may be biased to   head classes and tends to generate more head-class pseudo labels for unlabeled samples, leading to a more skewed degree of imbalance.}

{To address this issue, Distribution Alignment and Random Sampling (DARS)~\cite{he2021re}  proposed to  regard the label frequencies of labeled data as  a reference  and   enforce the     label frequencies of the generated pseudo labels to be consistent with the labeled ones. Instead of using training label frequencies,  Class-rebalancing Self-training (CReST)~\cite{wei2021crest}   found that  \emph{the precision of the supervised  model  on tail classes is surprisingly high}, and thus proposed to select more tail-class samples  for   online pseudo labeling  in each iteration, so that the re-trained model can obtain better performance on tail classes.
Beyond classification tasks, MosaicOS~\cite{zhang2021mosaicos} resorted to  other object-centric images  to boost long-tailed object detection. Specifically, it first pre-trains the model with labeled scene-centric images from  the original  detection dataset, and then  uses the pre-trained model to generate pseudo bounding boxes   for object-centric images, \eg ImageNet-1K~\cite{deng2009imagenet}. After that, MosaicOS fine-tunes the pre-trained model in two stages, \ie first fine-tuning with the pseudo-labeled object-centric images and then fine-tuning with the original labeled scene-centric images. In this way, MosaicOS    alleviates the negative influence of data discrepancies and effectively improves   long-tailed performance.}

{\textbf{Discussions}. These transfer learning methods  are complementary to each other, which brings additional information from different perspectives  to long-tailed learning. Most of them  can be used together for real applications if the resources permit and the combination pipeline is designed well. More concretely, when using model pre-training, the trade-off between supervised discrimination learning and self-supervised class-balanced learning should be tuned~\cite{kang2021exploring}, which contributes to better long-tailed learning performance. In addition, knowledge distillation with multi-experts  can usually achieve better performance than   distillation with a single teacher. In head-to-tail model transfer,  GIST is a better candidate than MetaModelNet due to its simplicity. Lastly, the use of self-training methods depends on   task requirements and what   unlabeled samples  are available at hand.}

\subsubsection{Data Augmentation}
{Data Augmentation aims to enhance the size and quality of  datasets by applying   pre-defined transformations to each data$/$feature for   model training~\cite{perez2017effectiveness,shorten2019survey}. In long-tailed learning, there are two types of  augmentation methods that have been explored, \ie transfer-based augmentation and non-transfer   augmentation.}


{\textbf{Head-to-tail transfer augmentation} seeks to transfer the knowledge from head classes to augment    tail-class samples. For example,  Major-to-Minor translation (M2m)~\cite{kim2020m2m} proposed to augment tail classes by translating head-class samples to tail-class ones via perturbation-based optimization, which is essentially similar to adversarial attack. The translated tail-class samples are used to  construct a more balanced training set for model training.}

{Besides the data-level transfer in M2m, most studies explore feature-level transfer. For instance, Feature Transfer Learning (FTL)~\cite{yin2019feature} found that \emph{tail-class samples have much smaller intra-class variance than head-class samples, leading to biased feature spaces and decision boundaries}. To address this, FTL exploits the knowledge of intra-class variance from head classes to guide feature augmentation for tail-class samples, so that the tail-class features have higher intra-class variance. 
Similarly, LEAP~\cite{liu2020deep}  constructs ``feature cloud” for each class, and  transfers the distribution knowledge of  head-class feature clouds to  enhance the intra-class variation of   tail-class feature clouds.  As a result,  the distortion of the intra-class feature variance among classes is  alleviated, leading to better tail-class performance.}

{Instead of using the intra-class variation information, Rare-class Sample Generator (RSG)~\cite{wang2021rsg} proposed to  dynamically estimate a set of feature centers for each class, and use  \emph{the feature displacement between head-class sample features and their nearest intra-class feature center} to augment each tail sample feature for enlarging the tail-class feature space.  Moreover, Online Feature Augmentation (OFA)~\cite{chu2020feature} proposed to use class activation maps~\cite{zhou2016learning} to decouple sample features into class-specific   and class-agnostic ones. Following that,  OFA  augments tail classes by combining the class-specific features of tail-class samples with   class-agnostic features from head-class samples.}


\textbf{Non-transfer  augmentation}   seeks to improve or design conventional data augmentation  methods to address long-tailed problems.  {SMOTE~\cite{han2005borderline}, a classic over-sampling method for non-deep  class imbalance, can   be applied to deep long-tailed problems to generate   tail-class samples by mixing several intra-class neighbouring samples. Recently, MiSLAS~\cite{zhong2021improving} further investigated  data mixup  in deep long-tailed learning}, and found that (1) \emph{data mixup helps to remedy model over-confidence}; (2) \emph{mixup has a positive effect on representation learning but a negative or negligible effect on classifier learning in the decoupled training scheme}~\cite{kang2019decoupling}. Following these observations, MiSLAS proposed to use   data mixup    to enhance representation learning in the  decoupled scheme.  In addition, Remix~\cite{chou2020remix} also resorted to data mixup for long-tailed learning and introduced a re-balanced mixup method to particularly enhance tail  classes. 

{Instead of using data mixup, FASA~\cite{zang2021fasa} proposed to generate new data features for each class,    based on class-wise Gaussian priors with their mean and variance estimated from previously observed samples. Here, FASA exploits the model classification loss on a balanced   validation set to adjust feature sampling rates   for different classes, so that the under-represented tail classes can be augmented more than head classes.
With a similar idea, Meta Semantic Augmentation (MetaSAug)~\cite{li2021metasaug} proposed to augment   tail classes with a variant of implicit semantic data augmentation (ISDA)~\cite{wang2019implicit}. Specifically,  ISDA estimates the class-conditional statistics (\ie covariance matrices from sample features) to obtain semantic directions, and generates diversified augmented samples by translating sample features along with diverse semantically meaningful directions. To better   estimate the covariance matrices for tail classes, MetaSAug explored meta learning to guide the learning of covariance matrices for each class with the class-balanced loss~\cite{cui2019class}, leading to more  informative synthetic  features.}  

{\textbf{Discussions}. Data augmentation based methods  attempt   to  address the class imbalance at the sample or feature levels.   The goals of these   methods are consistent, so they can be used simultaneously if the combination pipeline is constructed well. Among its two   subtypes, head-to-tail transfer augmentation is   more intuitive   than non-transfer augmentation. More specifically,    head-to-tail transfer at the feature level (\eg RSG) seems to perform better than transfer at the sample level (\eg M2m). In the feature-level transfer augmentation, RSG is preferred thanks to its easy-to-use source code, whereas  the intra-class variation in FTL and LEAP may be less informative for augmentation when the feature dimension is very high. In non-transfer augmentation, mixup-based strategies  are usually used thanks to their simplicity, where MiSLAS has demonstrated promising performance. In contrast, the   class-wise Gaussian priors in FASA and the covariance matrices in MetaSAug may be difficult to estimate  in various real scenarios.}

\subsubsection{Summary} 
{Information augmentation  addresses the long-tailed problems by introducing additional knowledge, and thus  is compatible with and complementary  to   other two method types, \ie class re-balancing and module improvement. For the same reason, its two method subtypes, \ie transfer learning and data augmentation, are also complementary to each other. More concretely, both the subtypes  are able to  improve tail-class performance without sacrificing head-class performance  if designed carefully. Considering  that   all classes are important in long-tailed learning, this type of  method   is   worth further exploring. Moreover, data augmentation  is a very fundamental technique and  can be used for a variety of long-tailed   problems, which makes it more practical than other paradigms in real-world applications. However, simply using existing \emph{class-agnostic} augmentation techniques for improving long-tailed learning is unfavorable, since they ignore the class imbalance and  inevitably augment more   head-class samples than tail-class samples. How to better conduct data augmentation for long-tailed learning is still an open question.}

\subsection{Module Improvement}
Besides   re-balancing and information augmentation,   researchers also explored methods to improve network modules in long-tailed learning.   These methods can be divided into four  categories: (1) representation learning improves the feature extractor; (2) classifier design enhances the model classifier; (3) decoupled training {aims to} boost the learning of both the feature extractor and the classifier;  (4) ensemble learning improves the  whole   architecture.

\subsubsection{Representation Learning}
{Existing  long-tailed learning    methods improve representation learning   based on three main paradigms, \ie  metric learning, prototype learning, and sequential training. }
 
{\textbf{Metric learning} aims at designing task-specific distance metrics for establishing similarity or dissimilarity between data.  In deep long-tailed learning, metric learning based methods seek  to explore various distance-based losses to learn a   discriminative  feature space for long-tailed data.   One example is Large Margin Local Embedding (LMLE)~\cite{huang2016learning}, which introduced a quintuplet loss to learn representations that maintain both inter-cluster and inter-class margins.   Unlike the triplet loss~\cite{hermans2017defense} that samples two contrastive pairs, LMLE presented a quintuplet sampler to sample four contrastive pairs, including a positive pair and three negative pairs. The positive pair is   the most distant intra-cluster sample, while the negative pairs include two inter-clusters samples from the same class (one is the nearest and one is the most distant within the same cluster) and the   nearest inter-class sample. Following that, LMLE introduced a  quintuplet loss to encourage the sampled quintuplet to follow a specific distance order. In this way, the learned representations preserve not only locality across intra-class clusters but also discrimination between classes. Moreover, each data batch    contains the same number of samples from different classes for class re-balancing.}   
{However, LMLE does not consider  the sample differences among head and tail classes. To address this,  Class Rectification Loss (CRL)~\cite{dong2017class} explored hard pair mining and proposed to construct more hard-pair triplets for tail classes, so that tail-class features  can have a larger degree of intra-class compactness and inter-class distances.}

{Rather than  sampling triplets  or quintuplets, range loss~\cite{zhang2017range} innovated representation learning by using the overall distances among all sample pairs within one mini-batch. In other words, the range loss uses statistics over the whole batch  and thus alleviates   the bias of data number imbalance over   classes. Specifically, range loss enlarges the inter-class distance by maximizing the distances of any two class centers within the mini-batch, and reduces the intra-class variation by minimizing the largest distances between intra-class samples.  In this way, the range loss   obtains features with better discriminative abilities and less imbalanced bias.}

{Recent studies also explored contrastive learning for long-tailed problems. KCL~\cite{kang2021exploring} proposed a $k$-positive contrastive loss to learn a balanced  feature space, which helps to alleviate  the class imbalance and improve model generalization. 
Parametric contrastive learning (PaCo)~\cite{cui2021parametric} further innovated supervised contrastive learning by adding a set of parametric learnable class centers, which   plays the same role as a classifier if regarding the class centers as the classifier weights.   Following that,
Hybrid~\cite{wang2021contrastive} introduced  a prototypical   contrastive learning strategy to enhance   long-tailed learning.   DRO-LT~\cite{samuel2021distributional} extended the  prototypical contrastive learning with distribution robust optimization~\cite{goh2010distributionally}, which makes the learned model more robust to   distribution shift.}

{\textbf{Prototype learning} based methods seek to learn class-specific feature prototypes to enhance long-tailed learning performance.  
Open Long-Tailed Recognition (OLTR)~\cite{liu2019large} innovatively explored the idea of feature prototypes to handle  long-tailed recognition in an open world, where the test set also includes open classes that do not appear in training data.   To address this task,   OLTR maintains a visual meta memory containing discriminative feature prototypes, and uses the features  sampled from the visual memory to augment the original features for better discrimination. Meanwhile, the sample features from novel classes are enforced to be far  away from the memory and closer to the origin point. In this way, the learned feature space enables OLTR to classify all seen classes  and detect novel classes. However, OLTR only maintains a  static  prototype memory and each class has only one prototype. Such a single prototype per class may fail to represent the real data distribution. To address this issue, Inflated Episodic Memory (IEM)~\cite{zhu2020inflated} further innovated the  meta-embedding memory by  a dynamical update scheme, in which each class has independent and differentiable memory blocks. Each memory block is updated to record the most discriminative feature  prototypes  of the corresponding categories, thus leading to better performance than OLTR.}

{\textbf{Sequential training} based methods learn data representation in a continual way.
For example, Hierarchical Feature Learning (HFL)~\cite{ouyang2016factors} took inspiration from  that each class has its individuality in discriminative visual representation. Therefore, HFL hierarchically clusters objects into visually similar class groups, forming a hierarchical cluster tree. In this cluster tree, the model in the original node is pre-trained on ImageNet-1K; the model in each child node  inherits the model parameters from its parent node and is then fine-tuned based on samples in the cluster node. In this way, the knowledge from the groups with massive classes is gradually transferred to their sub-groups with fewer classes.   
Similarly, Unequal-training~\cite{zhong2019unequal} proposed to divide the dataset into  head-class and tail-class subsets, and treat them differently in the training process. First, unequal-training   uses the head-class samples to train relatively discriminative and noise-resistant features with a new noise-resistant loss. After that, it uses tail-class samples to enhance the inter-class discrimination of representations via hard identity mining and a novel center-dispersed loss.  }


  

{\textbf{Discussions}. These representation learning methods seek to address the class imbalance at the feature level. The methods within each subtype are competing with each other (\eg KCL~\cite{kang2021exploring} vs PaCo~\cite{cui2021parametric} and OLTR~\cite{liu2019large}  vs IEM~\cite{zhu2020inflated}), while the methods from different subtypes may be complementary to each other (\eg KCL~\cite{kang2021exploring} and Unequal-training~\cite{zhong2019unequal}). Therefore, the pipeline must be carefully designed, if one wants to combine them together. Moreover, when handling real long-tailed applications, PaCo~\cite{cui2021parametric} is recommended to use thanks to its promising performance and open-source code. If there are open classes in test data, IEM~\cite{zhu2020inflated} is preferred. Other methods, like Unequal-training~\cite{zhong2019unequal}, can also be considered if they   suit   real scenarios.}  

\subsubsection{Classifier Design}\label{sec_classifier}
In addition to representation learning, researchers also explored different types of classifiers  to address   long-tailed problems. In generic visual   problems~\cite{he2016deep,he2020momentum},  the common practice of deep learning is to use linear classifier $p= \phi(w^{\top}f\small{+}b)$, where $\phi$ denotes the softmax function and  the bias term $b$ can be discarded.   
However,   long-tailed  class imbalance  often results in  larger classifier weight norms   for head classes than tail classes~\cite{yin2019feature}, which makes the linear classifier easily  biased  to dominant classes. 

         
      

{To address this, recent studies~\cite{liu2020deep,wu2021adversarial} proposed to use the scale-invariant cosine classifier $p = \phi((\frac{w^{\top}f}{\|w\|\|f\|})/\tau+ b)$, where  both the classifier weights  and sample features are normalized. Here, the temperature $\tau$ should be chosen reasonably~\cite{ye2020identifying}, or the classifier performance   would be negatively influenced. However, normalizing the feature space may harm its representation abilities. Therefore,  the $\tau$-normalized classifier~\cite{kang2019decoupling} rectifies the imbalance   by only adjusting the classifier weight norms  through a   $\tau$-normalization procedure. Formally, let $\tilde{w}= \frac{w}{\|w\|^{\tau}_2}$, where $\tau$ is   the  temperature factor for   normalization. When $\tau=1$, the $\tau$-normalization reduces to $L_2$ normalization, while when $\tau=0$, no scaling is imposed. Note that, the hyper-parameter $\tau$ can also be  trained with class-balanced sampling, and the resulting classifier is named the learnable weight scaling classifier~\cite{kang2019decoupling}.
Another approach to address classifier weight imbalance is to use the nearest class mean classifier~\cite{kang2019decoupling}, which first computes the  mean features for each class on the training set as the classifier, and then conducts prediction based on  the nearest neighbor algorithm~\cite{cover1967nearest}. }

{There are also some more complicated classifier designs based on hierarchical classification, causal inference or classifier knowledge transfer. For example, Realistic Taxonomic Classifier (RTC)~\cite{wu2020solving}  proposed to address  class imbalance with hierarchical classification by   mapping  images into a class taxonomic tree structure, where the hierarchy is defined by a set of classification nodes  and node relations.  Different samples are adaptively classified  at different hierarchical levels,} where the level at which the prediction is made depends on the sample classification difficulty and the classifier confidence.  Such a design favors correct decisions at intermediate levels rather than incorrect decisions at the leaves.

Causal classifier~\cite{tang2020long}   resorted to  causal inference for keeping the good and removing the bad momentum causal effects in long-tailed learning. The good causal effect indicates the beneficial factor that stabilizes gradients and accelerates training, while the bad causal effect indicates   the  accumulated long-tailed bias that leads to poor tail-class performance. To better approximate the bias information, the causal classifier applies a multi-head strategy to  divide the channel (or dimensions) of model weights and data features equally into $K$ groups. Formally, the causal classifier calculates the original logits by $p=\phi(\frac{\tau}{K}\sum_{k=1}^K \frac{(w^k)^{\top }f^k}{(\|w^k\|+\gamma)\|f^k\|})$, where  $\tau$ is the temperature factor and $\gamma$ is a hyper-parameter. This classifier is essentially the cosine classifier when $\gamma=0$. 
In inference, the causal classifier removes the bad causal effect by subtracting the prediction when the input is null, \ie $p=\phi(\frac{\tau}{K}\sum_{k=1}^K \frac{(w^k)^{\top}f^k}{(\|w^k\|+\gamma)\|f^k\|}-\alpha \frac{cos(x^k,\hat{d}^k)(w^k)^{\top}\hat{d}^k}{\|w^k\|+\gamma})$, 
where $\hat{d}$ is the unit vector of the exponential moving average features,   and $\alpha$ is a trade-off parameter to control the direct and indirect effects. More intuitively, the classifier records the bias  by computing the exponential moving average   features during training, and then removes the bad causal effect by subtracting the bias  from prediction logits during inference.

GIST classifier~\cite{liu2021gistnet} seeks to transfer the classifier geometric structure of head classes to tail classes. Specifically, the GIST classifier consists of a class-specific weight center (for encoding the class location) and a set of displacements (for encoding the class geometry). By exploiting the  relatively large displacements   from head classes to enhance   tail-class  weight centers, the GIST classifier is able to obtain better performance on tail classes.

\begin{figure*}  
\vspace{-0.15in}
  \begin{minipage}{0.32\linewidth}
   \centerline{\includegraphics[height=4cm]{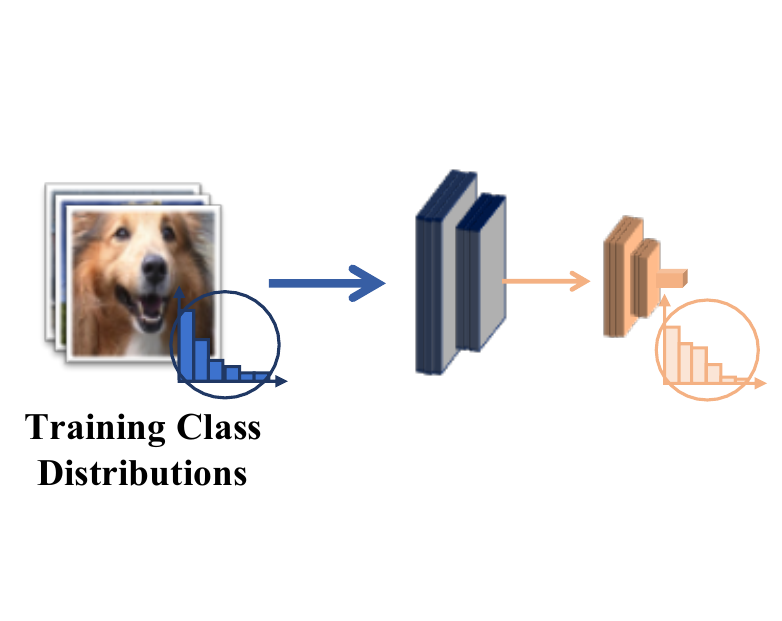}}
   \centerline{\small{(a) Standard training}}
  \end{minipage}
    \hfill
  \begin{minipage}{0.32\linewidth}
   \centerline{\includegraphics[height=4cm]{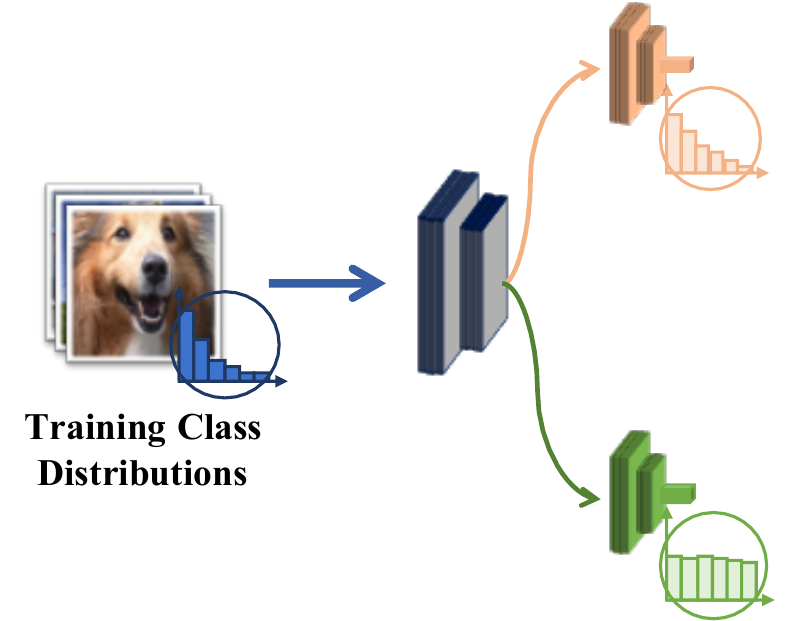}}
   \centerline{\small{(b) BBN~\cite{zhou2020bbn}, TLML~\cite{guo2021long}, SimCAL~\cite{wang2020devil}}}
  \end{minipage}
  \hfill
\begin{minipage}{0.32\linewidth}
   \centerline{\includegraphics[height=4cm]{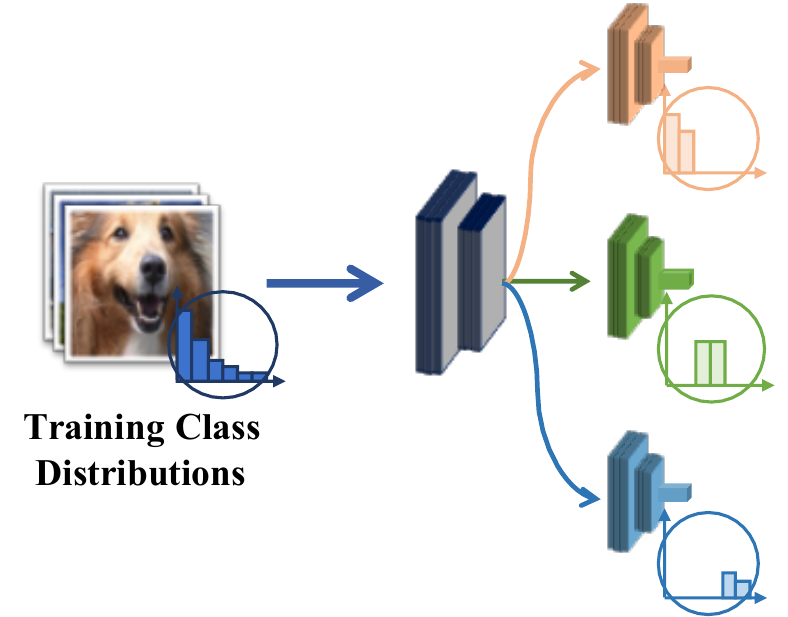}}
   \centerline{\small{(c) BAGS~\cite{li2020overcoming}, LFME~\cite{xiang2020learning}}}
  \end{minipage}
  \vfill 
  \vspace{0.1in}
  \vfill    
    \begin{minipage}{0.32\linewidth}
   \centerline{\includegraphics[height=4cm]{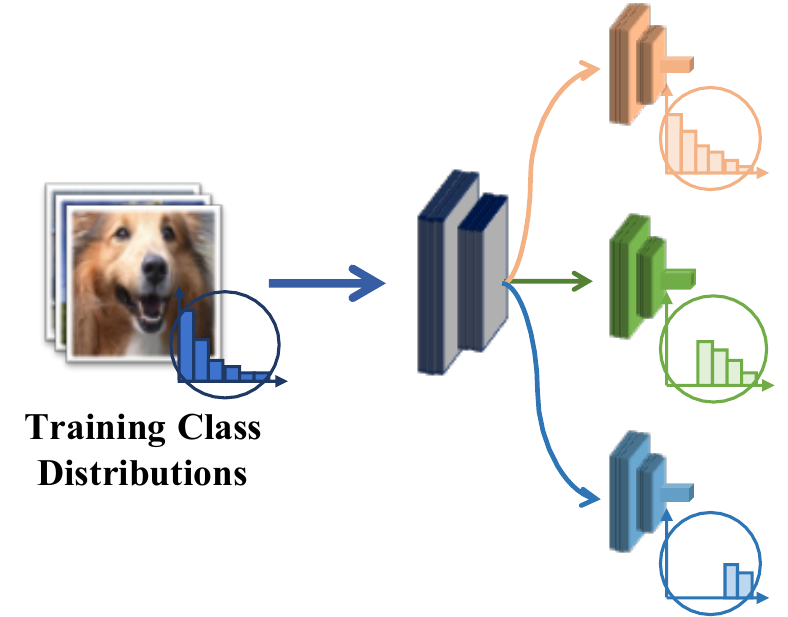}}
   \centerline{\small{(d) ACE~\cite{cai2021ace}, ResLT~\cite{cui2021reslt}}}
  \end{minipage}
  \hfill
  \begin{minipage}{0.32\linewidth}
   \centerline{\includegraphics[height=4cm]{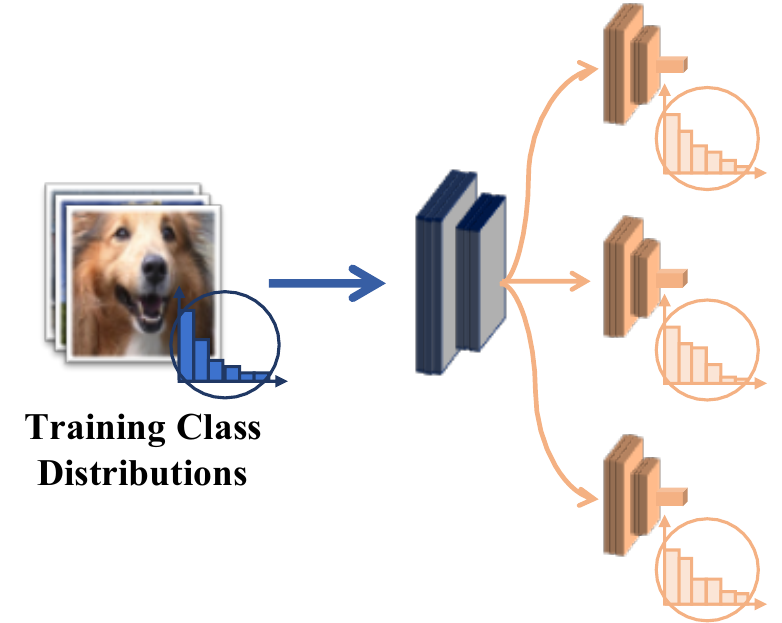}}
   \centerline{\small{(e) RIDE~\cite{wang2020long}}}
  \end{minipage}
  \hfill
  \begin{minipage}{0.32\linewidth}
   \centerline{\includegraphics[height=4cm]{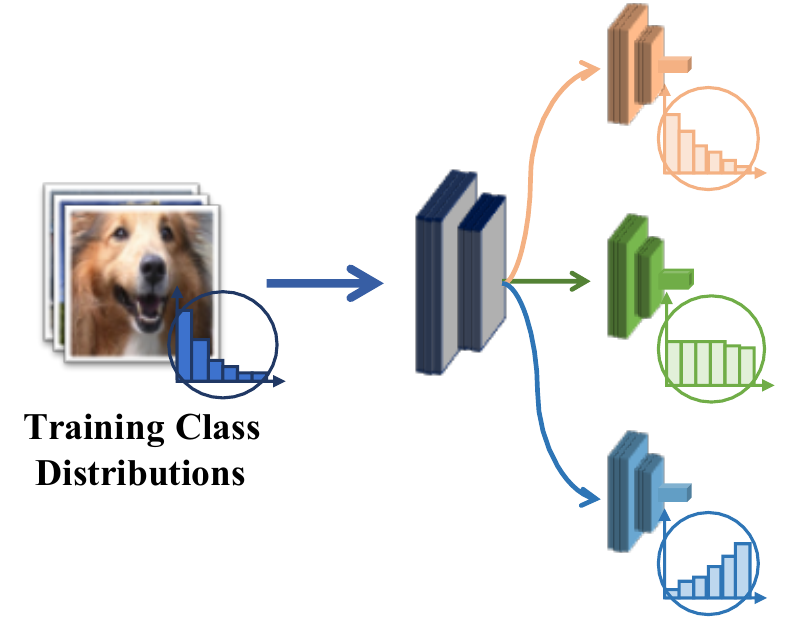}} 
   \centerline{\small{(f) SADE~\cite{zhang2021test}}}
  \end{minipage}
 \caption{Illustrations of existing ensemble-based  long-tailed   methods. {Compared to standard training (a), the trained  experts by ensemble-based methods (b-f) may have different expertise, \eg being skilled in different class distributions or different class subsets (indicated by different colors). For example, BBN and SimCAL train two experts for simulating the original long-tailed and uniform distributions so that they can address the two distributions well.  BAGS, LFME, ACE, and ResLT train multiple experts by sampling class subsets, so that different experts can particularly handle different sets of classes. SADE directly trains multiple experts to separately simulate long-tailed, uniform and inverse long-tailed class distributions from a stationary long-tailed   distribution, which enables it to handle   test sets with agnostic class distributions based on self-supervised aggregation.}}\label{fig_ensemble}  
\vspace{-0.1in}
\end{figure*}

{\textbf{Discussions}. These methods address the   imbalance at the classifier level. Note that these classifiers are exclusive to each other, and the choice of   classifiers also influences   other long-tailed methods. For example, the effects of data mixup are different for the linear classifier and the cosine classifier. Hence, when exploring new long-tailed approaches, it is better to first determine which classifier is used. Generally, the cosine classifier or the learnable weight-scaling  classifier are recommended, as they are empirically robust to the   imbalance and also easy to use. Moreover,  when designing feature prototype-based methods, the nearest class mean classifier is a good choice. More complicated classifier designs (\eg RTC, Causal and GIST) can also be considered if  real applications are  complex and hard to handle.}


\subsubsection{Decoupled Training}\label{sec_decoupled}
Decoupled training decouples the learning procedure into representation learning    and classifier training. {Here, decoupled training represents a general paradigm  for long-tailed learning instead of a specific approach. 
Decoupling~\cite{kang2019decoupling} was the pioneering work to introduce such a two-stage decoupled  training scheme.   It empirically evaluated different sampling strategies  (mentioned in Section~\ref{sec_resampling}) for representation learning in the first stage}, and then evaluated different classifier training schemes by fixing the  trained feature extractor in the second stage.  In the classifier learning stage, there are also four methods, including classifier re-training  with class-balanced sampling, the nearest class mean classifier, the $\tau$-normalized classifier, and {the learnable weight-scaling classifier. The main observations are twofold: (1) \emph{random sampling is surprisingly the best strategy for representation learning} in decoupled training; (2)  \emph{re-adjusting the classifier leads to significant performance improvement} in long-tailed recognition.}

{Following this scheme, KCL~\cite{kang2021exploring} empirically observed that \emph{a balanced feature space is beneficial to long-tailed learning}. Therefore, it innovated the decoupled training scheme by    developing a $k$-positive contrastive loss to learn a more class-balanced and class-discriminative feature space, which leads to   better long-tailed learning performance.  Moreover, MiSLAS~\cite{zhong2021improving} empirically observed  that \emph{data mixup is beneficial to features learning but has a negative/negligible effect on classifier training under the two-stage decoupled training scheme}. Therefore, MiSLAS proposed to enhance the representation learning  with data mixup in the first stage, while  applying a  label-aware smoothing strategy for better classifier generalization in the second stage.}

{Several recent     studies  particularly enhanced the classifier training  stage.
For example, OFA~\cite{chu2020feature} innovated the classifier re-training through     tail-class feature augmentation.  
SimCal~\cite{wang2020devil}  enhanced the  classifier training stage by calibrating the classification head with a novel bi-level class-balanced sampling strategy for  long-tailed instance segmentation. DisAlign~\cite{zhang2021distribution} innovated the     classifier training with a new adaptive logit adjustment  strategy.   
Very recently, DT2~\cite{Desai2021Learning} applied the scheme of decoupled training to the scene graph generation task, which demonstrates  the effectiveness of decoupled training in handling long-tailed  visual relation learning.}

{\textbf{Discussions}. Decoupled training methods resolve the class imbalance issue at both the feature and classifier levels.
Under ideal conditions, combining different  methods  can lead to better long-tailed performance, \eg using self-supervised pre-training~\cite{kang2021exploring} and mixup augmentation~\cite{zhong2021improving} together for better representation learning, and applying label-aware smoothing~\cite{zhong2021improving} and tail-class feature augmentation~\cite{chu2020feature} together for better classifier tuning. Therefore, it is recommended to use MiSLAS~\cite{zhong2021improving} as a base method and use different tricks on it. Note that some representation  methods are also competing to each other, \eg different sampling methods for representation learning~\cite{kang2019decoupling}.} 

{The classifier learning stage   does not introduce too many computation costs but can lead to significant performance gains. This makes decoupled training attract increasing attention. One critique is that the accumulated training stages make decoupled training less practical to be integrated with existing well-formulated methods for other long-tailed problems like object detection and instance segmentation. Despite this, the idea of decoupled training is conceptually simple  and thus can be easily used to design new methods for resolving  various  long-tailed    problems, like DT2~\cite{Desai2021Learning}.}

\subsubsection{Ensemble Learning}
Ensemble learning based methods strategically generate  and combine multiple network modules (namely, multiple experts) to solve long-tailed visual learning problems. We summarize the main schemes of existing ensemble-based  methods in Fig.~\ref{fig_ensemble}, which will be detailed  as follows. 
 
{BBN~\cite{zhou2020bbn} proposed to use two network branches, \ie a conventional learning branch and a re-balancing branch  (cf.~Table~\ref{fig_ensemble}(b)),  to handle long-tailed recognition. To be specific, the conventional learning branch applies uniform sampling to simulate the original long-tailed training distribution, while the re-balancing branch applies a reversed sampler to sample more tail-class samples in each mini-batch for 
improving tail-class performance. The predictions of two branches are dynamically combined during training, so that    the learning focus of BBN gradually changes from head classes to 
tail classes.
Following BBN, LTML~\cite{guo2021long}   applied the   bilateral-branch network scheme to solve  long-tailed multi-label classification. To be specific, LTML trains each branch using the sigmoid cross-entropy loss for multi-label classification  and enforces a logit consistency loss to improve the consistency of the two branches. 
Similarly, SimCal~\cite{wang2020devil}   explored a dual classification head scheme, a conventional classification head and a calibrated classification head, to address long-tail instance segmentation.  Based on a  new bi-level sampling strategy, the calibrated classification head is able to improve the performance on tail classes, while  the original  head aims to maintain the performance on head classes.}

{Instead of bilateral branches,  BAGS~\cite{li2020overcoming} explored a multi-head scheme to address long-tailed object detection. Specifically, BAGS took inspiration from an observation that learning a more uniform distribution with fewer samples is sometimes easier than learning a long-tailed distribution with more samples. Therefore, BAGS    divides classes into several  groups, where the classes in each  group have a similar number of training data. Then, BAGS applies multiple classification heads  for prediction, where different   heads are trained on different class   groups  (cf.~Table~\ref{fig_ensemble}(c)). In this way, each  classification head performs the softmax operation on classes with a similar number of training data,   thus avoiding the negative influence of class imbalance. Moreover, BAGS also introduces a label of “other classes”  into each group to alleviate the contradiction among different heads.
Similar to BAGS, LFME~\cite{xiang2020learning} divides the long-tailed dataset into several subsets with  smaller  class imbalance degrees, and trains multiple experts with different sample subsets. Based on these experts, LFME then learns a unified student model using adaptive knowledge distillation from   multiple teachers.} 

Instead of  division into several balanced sub-groups, ACE~\cite{cai2021ace}  divides classes into several skill-diverse subsets:  one   subset contains all classes; one contains middle and tail classes;  another one has only tail classes (cf.~Table~\ref{fig_ensemble}(d)). ACE then trains  multiple experts with various class subsets, so that different experts have specific and complementary skills. Moreover, considering  that various subsets have   different sample numbers, ACE also applies a distributed-adaptive optimizer to adjust the learning rate for different experts. A similar idea of ACE was also explored in ResLT~\cite{cui2021reslt}.

{Instead of  dividing the dataset, RIDE~\cite{wang2020long} uses    all training samples  to  train multiple experts  with softmax loss respectively (cf.~Table~\ref{fig_ensemble}(e)),    and enforces a KL-divergence based  loss to improve the diversity among various experts. Following that, RIDE applies an expert assignment module to improve computing efficiency. Note that training each expert with the softmax loss independently boosts the ensemble  performance on long-tailed learning a lot. However, the learned experts by RIDE are not diverse enough. }
  
{Self-supervised Aggregation of Diverse Experts (SADE)~\cite{zhang2021test}  explored a new multi-expert scheme to handle  test-agnostic long-tailed recognition, where the test class distribution can be either uniform, long-tailed or even inversely long-tailed. To be specific, SADE developed a novel spectrum-spanned multi-expert framework (cf.~Table~\ref{fig_ensemble}(f)), and innovated the expert training scheme by introducing   diversity-promoting expertise-guided losses that train different experts to handle different class distributions, respectively. In this way, the learned experts are more diverse than RIDE, leading to better ensemble performance, and integratedly span a wide spectrum of possible class distributions. In light of this, SADE further introduced a self-supervised learning method, namely prediction stability maximization, to adaptively aggregate experts at test time for better handling unknown test class distribution. }

{\textbf{Discussions}. These ensemble-based methods address the class imbalance at the model level. As they require particular manners for multi-model design and training (cf.  Fig.~\ref{fig_ensemble}), they are exclusive to each other and usually cannot be used together. More specifically, the methods with bilateral branches like BBN and TLML only have two experts, whose empirical performance has been shown worse than the approaches with more experts. Moreover, the methods with experts trained on  class subsets like BAGS and ACE   may suffer from  expert inconsistency in terms of different   label spaces, which makes the aggregation of  experts  difficult and may lead to poor performance in real applications. Instead, RIDE trains multiple experts with all samples but the resulting multiple experts are not diverse enough. In contrast, SADE is able to train skill-diverse experts with the same label space, and thus is recommended for real applications.  One concern of  these ensemble-based methods is that they generally lead to higher computational costs due to the use of multiple experts. Such a concern, however, can be alleviated by using a shared feature extractor. Moreover,   efficiency-oriented expert assignment  and knowledge distillation strategies~\cite{xiang2020learning,wang2020long} can also reduce computational complexity.}

\subsubsection{Summary}  
{Module improvement based methods seek  to address long-tailed  problems by improving network modules. Specifically,  representation learning and classifier design are fundamental problems of deep  learning, being worth further  exploring for long-tailed problems. Both representation learning and classifier design  are complementary to decoupled training. The scheme of decoupled training  is conceptually simple  and   can be easily used to design new methods for resolving real long-tailed applications.  In addition, ensemble-based methods, thanks to the aggregation of multiple experts,  are able to  achieve better long-tailed performance without sacrificing   the performance on any class subsets, \eg  head   classes. Since all classes are important, such a superiority enables   ensemble-based methods to be  a better choice for real applications compared to existing class re-balancing  methods that usually improve tail-class performance  at the cost of  lower head-class performance. Here, both ensemble-based methods  and decoupled training   require specific model training and design manners, so it is not easy to use them together unless very careful design.}

{Note that most module improvement methods are developed based on  fundamental class re-balancing  methods. Moreover, module improvement methods are complementary to information augmentation methods. Using them together can usually achieve better performance  for real-world long-tailed applications.}


\section{Empirical Studies}\label{Sec4}
This section empirically analyzes  existing  long-tailed learning methods. To begin with, we introduce a new evaluation metric.

\subsection{Novel Evaluation Metric}
The key goal of long-tailed learning is to  handle the class imbalance for better model performance. 
Therefore, the common evaluation protocol~\cite{cao2020domain,kang2021exploring} is directly using the top-1 test accuracy (denoted by $A_t$) to judge how well long-tailed methods perform and  which method   handles class imbalance better. Such a metric, however, cannot accurately reflect the  relative superiority among different methods when handling class imbalance, as the top-1 accuracy is also influenced by other factors apart from class imbalance. For example,  long-tailed   methods like ensemble learning (or data augmentation)  also improve the performance of models, trained on  a  balanced training set. In such cases, it is hard to 
tell if the performance gain is from the alleviation of class imbalance or from better network architectures (or more data information).

To  better evaluate the method  effectiveness  in handling class imbalance, we explore a new metric, namely \textbf{relative accuracy} $A_r$, to alleviate the influence of   unnecessary factors in long-tailed learning. To this end, we first compute an empirically upper reference accuracy $A_u = \max (A_v, A_b)$, which is the maximal value between the  \emph{vanilla accuracy} $A_v$ of the   backbone trained on a  balanced training set with cross-entropy and the  \emph{balanced accuracy} $A_b$ of the model  trained on  a balanced training set with the corresponding long-tailed method. {Here, the balanced training set is \emph{a variant of the long-tailed training set, where the total  data number is similar but each class  has the same  
number of data}}. This upper reference accuracy, obtained from the balanced training set, is used to alleviate the   influence  apart from class imbalance, and  then the \emph{relative accuracy} is 
defined by $A_r=\frac{A_t}{A_u}$. 
{Note that this metric is mainly designed for empirical understanding, \ie to evaluate to what extent existing methods resolve the class imbalance. We conduct this analysis based on the ImageNet-LT dataset~\cite{liu2019large}, where a corresponding   balanced training set variant can be built by  sampling from the original ImageNet following~\cite{kang2021exploring}.  }

\subsection{Experimental Settings}
We then introduce the  experimental settings.

\textbf{Datasets}.
We adopt the widely-used ImageNet-LT~\cite{liu2019large} {and iNaturalist 2018~\cite{van2018inaturalist}} as the benchmark long-tailed dataset  for empirical studies. Their  dataset statistics  can be found in Table~\ref{dataset}. Besides the performance regarding all classes,  we also report performance on three class subsets: Head (more than 100 images), Middle  (20$\sim$100 images) and Tail (less than 20 images).  

\textbf{Baselines}.
We select long-tailed    methods via   two criteria: (1)  the  source codes are publicly available or easy to re-implement; (2)  the methods are evaluated  on ImageNet-LT in the corresponding papers. As a result,  more than 20  methods are   empirically evaluated in this paper, including baseline (\textbf{Softmax}), 
class-sensitive learning  (\textbf{Weighted Softmax}, \textbf{Focal loss}~\cite{lin2017focal}, \textbf{LDAM}~\cite{cao2019learning}, \textbf{ESQL}~\cite{tan2020equalization}, \textbf{Balanced Softmax}~\cite{jiawei2020balanced}, \textbf{LADE}~\cite{hong2020disentangling}),  
logit adjustment  (\textbf{UNO-IC}~\cite{tian2020posterior}),   
transfer learning  (\textbf{SSP}~\cite{yang2020rethinking}),  
data augmentation  (\textbf{RSG}~\cite{wang2021rsg})
representation learning  (\textbf{OLTR}~\cite{liu2019large}, \textbf{PaCo}~\cite{cui2021parametric}).
classifier design  (\textbf{De-confound}~\cite{tang2020long}),  
decoupled training  (\textbf{Decouple-IB-CRT}~\cite{kang2019decoupling}, \textbf{CB-CRT}~\cite{kang2019decoupling}, \textbf{SR-CRT}~\cite{kang2019decoupling}, \textbf{PB-CRT}~\cite{kang2019decoupling},  \textbf{MiSLAS}~\cite{zhong2021improving}),
ensemble learning  (\textbf{BBN}~\cite{zhou2020bbn}, \textbf{LFME}~\cite{xiang2020learning}, \textbf{RIDE}~\cite{wang2020long}, \textbf{ResLT}~\cite{cui2021reslt}, \textbf{SADE}~\cite{zhang2021test}). 

\textbf{Implementation details}.
We implement all experiments  in PyTorch. Following~\cite{hong2020disentangling,wang2020long,kang2019decoupling}, {we use ResNeXt-50 for ImageNet-LT and and ResNet-50 for iNaturalist 2018 as  the network backbones}  for all methods. We conduct model training with the SGD optimizer based on batch size 256, momentum 0.9 and weight decay factor 0.0005, and  learning rate 0.1 (linear LR decay). For method-related hyper-parameters, we set the values   by either directly following the original papers or  manual tuning if the default values perform poorly. Moreover, we use the same  basic data augmentation (\ie random resize and crop to 224, random horizontal flip, color jitter, and normalization) for all methods.

\subsection{{Results on  ImageNet-LT}}

\begin{figure}   
  \begin{minipage}{1\linewidth}
   \centerline{\includegraphics[height=5.3cm]{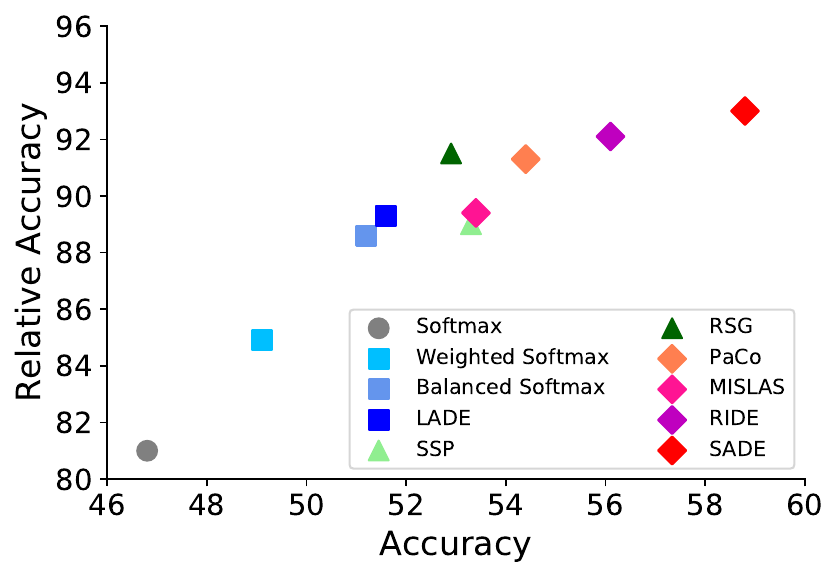}} 
  \end{minipage}
 \vspace{-0.1in}
 \caption{Performance trends of long-tailed learning  methods in terms of accuracy and relative accuracy under 200  epochs.
 Here, the shape of $\circ$ indicates the softmax baseline; $\square$ indicates class re-balancing; $\bigtriangleup$ and $\diamondsuit$ are information augmentation and module improvement methods, respectively. Different colors  represent different methods. }\label{fig_trend}  
 \vspace{-0.1in}
\end{figure}

\textbf{{Observations on all classes.}}  Table~\ref{table_performance} and Fig.~\ref{fig_trend} report  the average performance of ImageNet-LT over all classes. From these results, we have several observations on the overall method progress and different method types. As shown in Table~\ref{table_performance}, almost all long-tailed methods perform better than the Softmax baseline in terms of accuracy, which demonstrates the effectiveness of long-tailed learning. Even so, there are two methods performing slightly worse than Softmax, \ie Decouple-CB-CRT~\cite{kang2019decoupling} and BBN~\cite{zhou2020bbn}. We speculate that the poor performance of Decouple-CB-CRT results from poor representation learning by class-balanced sampling in the first stage of decoupled training (refer to~\cite{kang2019decoupling} for more empirical observations). The poor results of BBN (based on the official codes) may come from the cumulative learning strategy, which gradually adjusts the learning focus from head classes to tail classes; at the end of the training, however,  it may put too much focus on the tail ones. As a result, despite the better tail-class performance, the model accuracy on head classes drops significantly (c.f.   Table~\ref{table_setperformance}), leading to worse average performance.

In addition to accuracy, we also evaluate long-tailed  methods based on upper reference accuracy (UA) and relative accuracy (RA). Table~\ref{table_performance} shows that most methods have the same UA  as the baseline model, but there are still some methods having higher  UA, \eg SSP, MiSLAS, and SADE. For these methods, the performance improvement  comes  not only  from the alleviation of class imbalance, but also from other factors, like data augmentation or better network architectures. Therefore, simply using accuracy for evaluation is not comprehensive enough, while  the proposed RA metric provides a good complement as it alleviates the influences of factors apart from class imbalance. For example, MiSLAS, based on data mixup, has higher accuracy than Balanced Softmax under 90 training epochs, but it also has higher UA. As a result, the relative accuracy of  MiSLAS is lower than Balanced  Softmax, which means that Balanced  Softmax alleviates class imbalance better than MiSLAS under 90 training epochs. 

Although some recent high-accuracy methods   have  lower RA, the overall development trend of long-tailed learning is still positive, as shown in  Fig.~\ref{fig_trend}. Such a performance trend demonstrates that recent studies of long-tailed learning  make real progress. Moreover, the RA of  the state-of-the-art SADE is 93.0, which implies that there is still room for improvement in the future.

 \begin{table*}[t]  
  \begin{center}   
  \begin{minipage}{0.5\linewidth}
 \caption{Results on ImageNet-LT   in terms of accuracy (Acc), upper reference accuracy~(UA),  relative accuracy~(RA) under 90 or 200 training epochs. In this table, CR, IA and MI indicate  class re-balancing, information augmentation and  module improvement, respectively.}\label{table_performance}  
 \vspace{-0.1in}
    \begin{center} 
    \scalebox{0.85}{  
    \begin{threeparttable} 
	\begin{tabular}{llccccccc}\toprule
        \multirow{2}{*}{Type} & \multirow{2}{*}{Method} &\multicolumn{3}{c}{90 epochs}&&\multicolumn{3}{c}{200 epochs} \cr\cmidrule{3-5}  \cmidrule{7-9}  
        & & Acc  & UA  & RA  && Acc  & UA  & RA      \cr
        \midrule
          Baseline & Softmax	& 45.5 & 57.3  & 79.4 &&46.8 	& 57.8 & 81.0  \\\midrule
        \multirow{7}{*}{CR}  &Weighted Softmax& 47.9 & 57.3 & 83.6  && 49.1 &57.8   & 84.9\\

        & Focal loss~\cite{lin2017focal}& 45.8 & 57.3  & 79.9 && 47.2	& 57.8 & 81.7  \\
         &LDAM~\cite{cao2019learning} & 51.1 & 57.3  & 89.2 &&  51.1	& 57.8 & 88.4  \\ 
        & ESQL~\cite{tan2020equalization}& 47.3 & 57.3  & 82.5 && 48.0	& 57.8 & 83.0 \\
        & UNO-IC~\cite{tian2020posterior}& 45.7 & 57.3  & 81.4 && 46.8	& 58.6 & 79.9 \\
         &Balanced Softmax~\cite{jiawei2020balanced}& 50.8 & 57.3  & 88.7 &&  51.2	& 57.8 & 88.6 \\
         & LADE~\cite{hong2020disentangling}  & 51.5 & 57.8  & 89.1 && 51.6	& 57.8 & 89.3 \\
        \midrule

       \multirow{2}{*}{IA} & SSP~\cite{yang2020rethinking}& 53.1 & 59.6  & 89.1  && 53.3 	& 59.9 & 89.0  \\
        & RSG~\cite{wang2021rsg}   & 49.6 & 57.3  & 86.7 && 52.9	& 57.8 & 91.5 \\
        \midrule
       
        \multirow{13}{*}{MI} & OLTR~\cite{liu2019large}& 46.7 & 57.3  & 81.5 && 48.0	& 58.4 & 82.2  \\
        & PaCo~\cite{cui2021parametric} & 52.7 & 58.7  & 89.9  && 54.4 	& 59.6 &  91.3 \\ 
        & De-confound~\cite{tang2020long}& 51.8 & 57.7  & 89.8 &&  51.3	& 57.8 & 88.8 \\
        \cmidrule{2-9} 
        & Decouple-IB-CRT~\cite{kang2019decoupling}  & 49.9 & 57.3  & 87.1 && 50.3	& 58.1 &86.6  \\
        &Decouple-CB-CRT~\cite{kang2019decoupling}  & 44.9 & 57.3  & 78.4 && 43.0	& 57.8 & 74.4 \\
        &Decouple-SR-CRT~\cite{kang2019decoupling}  & 49.3 & 57.3  & 86.0 && 48.5	& 57.8 & 83.9  \\
        &Decouple-PB-CRT~\cite{kang2019decoupling}  & 48.4 & 57.3  & 84.5 && 48.1	& 57.8 & 83.2  \\
        &MiSLAS~\cite{zhong2021improving}  & 51.4 & 58.3  & 88.2 && 53.4	& 59.7 & 89.4 \\ 
         \cmidrule{2-9} 
        
        &BBN~\cite{zhou2020bbn}& 41.2 & 57.3  & 71.9 && 44.7	& 57.8 & 77.3  \\
        &LFME~\cite{xiang2020learning} & 47.0 & 57.3  & 82.0 && 48.0	& 57.8 & 83.0 \\ 
        &ResLT~\cite{cui2021reslt}& 51.6 & 57.3  & 90.1  && 53.2 	& 58.1 & 91.6 \\
        &RIDE~\cite{wang2020long} & 55.5 & 60.2  & 92.2 && 56.1	& 60.9 & 92.1 \\  
        & SADE~\cite{zhang2021test}& \textbf{57.3} & \textbf{61.9}  & \textbf{92.6}   && \textbf{58.8}	& \textbf{63.2} &  \textbf{93.0} \\
        
        \bottomrule
	\end{tabular}  
    \end{threeparttable}}
    \end{center}  
 \vspace{-0.1in}
 \end{minipage} 
\hfill  
 \begin{minipage}{0.48\linewidth}
    \caption{Accuracy results on ImageNet-LT  regarding head, middle and tail classes under 90 or 200 training epochs. In this table, WS indicates weighed softmax and BS indicates balanced softmax. The types of methods  are the same to Table~\ref{table_performance}.} \label{table_setperformance}    \vspace{-0.08in}
    \begin{center} 
    \scalebox{0.85}{  
    \begin{threeparttable} 
	\begin{tabular}{lccccccc}\toprule
        \multirow{2}{*}{Method} &\multicolumn{3}{c}{90 epochs}&&\multicolumn{3}{c}{200 epochs} \cr\cmidrule{2-4}  \cmidrule{6-8}  
        &  Head  & Middle  & Tail  && Head  & Middle  & Tail     \cr
        \midrule
        Softmax	& 66.5 & 39.0 &8.6  && 66.9	& 40.4 &12.6 \\ \midrule
        WS& 66.3  & 42.2 & 15.6   && 57.9 & 46.2   & 34.0  \\
        Focal loss~\cite{lin2017focal}& 66.9 & 39.2  & 9.2 && 67.0	& 41.0  & 13.1  \\ 
        LDAM~\cite{cao2019learning} & 62.3 & 47.4  & 32.5 &&  60.0	& 49.2 & 31.9  \\
        ESQL~\cite{tan2020equalization}& 62.5 & 44.0  & 15.7 && 63.1	& 44.6 & 17.2 \\
        UNO-IC~\cite{tian2020posterior}& 66.3 & 38.7  & 9.3 && 67.0	& 40.3 & 12.7 \\
        BS~\cite{jiawei2020balanced}& 61.7 & 48.0  & 29.9 &&  62.4	& 47.7 & 32.1 \\
        LADE~\cite{hong2020disentangling}  & 62.2 & 48.6  & 31.8 && 63.1	& 47.7 & 32.7 \\
        \midrule
         
        SSP~\cite{yang2020rethinking}& 65.6& 49.6  & 30.3  && 67.3 	& 49.1 & 28.3  \\
        RSG~\cite{wang2021rsg}   &\textbf{68.7} & 43.7  & 16.2 && 65.0	& 49.4 & 31.1 \\
        \midrule
         
        OLTR~\cite{liu2019large}& 58.2  & 45.5  & 19.5 && 62.9	& 44.6 & 18.8  \\
        PaCo~\cite{cui2021parametric} & 59.7 & 51.7  & 36.6  && 63.2 	& 51.6 &  39.2 \\
        De-confound~\cite{tang2020long}& 63.0 & 48.5  & 31.4 &&  64.9	& 46.9 & 28.1 \\
        \midrule
        IB-CRT~\cite{kang2019decoupling}  & 62.6 & 46.2  & 26.7 && 64.2	& 46.1 &26.0  \\
        CB-CRT~\cite{kang2019decoupling}  & 62.4 & 39.3  & 14.9 && 60.9	& 36.9 & 13.5 \\
        SR-CRT~\cite{kang2019decoupling}  & 64.1 & 43.9  & 19.5 && 66.0	& 42.3 & 18.0 \\
        PB-CRT~\cite{kang2019decoupling}  & 63.9 & 45.0  & 23.2 && 64.9	& 43.1 & 20.6  \\
        MiSLAS~\cite{zhong2021improving}  & 62.1 & 48.9  & 32.6 && 65.3	& 50.6 & 33.0 \\ 
         \midrule
        
        BBN~\cite{zhou2020bbn}& 40.0 & 43.3  & 40.8 && 43.3	& 45.9 & \textbf{43.7}  \\
        LFME~\cite{xiang2020learning} & 60.6 & 43.5  & 22.0 && 64.1	& 42.3 & 22.8 \\
        ResLT~\cite{cui2021reslt}& 57.8 & 50.4  & 40.0  && 61.6 	& 51.4 & 38.8 \\
        RIDE~\cite{wang2020long} & 66.9 & 52.3  & 34.5 && \textbf{67.9}	& 52.3 & 36.0 \\
        SADE~\cite{zhang2021test}& 65.3 & \textbf{55.2}  & \textbf{42.0}  && 67.2	& \textbf{55.3} &  40.0 \\
        \bottomrule
	\end{tabular}  
    \end{threeparttable}}
    \end{center} 
  \end{minipage}
  \end{center} 
  \vspace{-0.1in}
\end{table*}

We also evaluate the influence of different training epochs (\ie 90 and 200)   in Table~\ref{table_performance}. Overall, training with 200 epochs   leads to better performance for most long-tailed methods, because sufficient  training enables deep models to fit data better and learn better visual representations. However,  there are also some methods that perform better when only training 90 epochs, \eg De-confound and Decouple-CB-CRT. We speculate that, for  these methods,  90 epochs are enough  to train models well, while training more epochs does not bring additional benefits but increases the training difficulties  
since it also influences the learning rate decay scheme.

\textbf{Observations on different method types.} {We next analyze different method types in   Table~\ref{table_performance}. To begin with, almost all class re-balancing (CR) methods all beneficial to long-tailed learning performance, compared to the baseline model. Among them,   LADE, Balanced Softmax and LDAM achieve  state-of-the-art.
Moreover, Focal loss was initially proposed to handle  object detection~\cite{lin2017focal}. However, when handling a highly large number of   classes (\eg 1,000 in ImageNet-LT), Focal loss cannot perform well and only leads to  marginal improvement.  In LDAM, there is a deferred re-balancing optimization schedule in addition to the LDAM loss. Simply learning with the LDAM loss without the deferred scheme cannot achieve promising results. In addition, as shown in Table~\ref{table_performance}, the upper reference accuracy of most class-sensitive   methods is the same, so their relative accuracy is positively correlated to accuracy. Hence, the accuracy improvement in this method type can accurately reflect the alleviation of class imbalance.}

{In information augmentation (IA), both SSP  (transfer learning) and  RSG  (data augmentation) help to handle long-tailed   imbalance.  Although  SSP also improves upper reference accuracy, its relative accuracy is increased more significantly,   implying that  the performance gain   mostly comes from   handling the class imbalance. 
In  module improvement (MI), all  methods contribute to addressing the   imbalance. By now, the state of the art   is  ensemble-based long-tailed methods, \ie SADE  and RIDE, in terms of both accuracy and relative accuracy. Although ensemble learning   also improves upper reference accuracy, the performance gain from  handling imbalance is more significant, leading to higher  relative accuracy.}

\textbf{{Results on different class subsets}}.
{We then report the  performance in terms of  different  class subsets. As shown in Table~\ref{table_setperformance}, almost all methods improve   tail-class and middle-class performance at the cost of lower head-class performance.} The head classes, however, are also important in long-tailed learning, so it is necessary to  improve long-tailed performance without sacrificing the performance of the head. Potential solutions include information augmentation and ensemble learning, \eg SSP  and SADE. 
By comparing both Tables~\ref{table_performance} and~\ref{table_setperformance}, one can find that the overall performance gain   largely depends on the improvement of middle  and tail classes; hence, how to improve their performance is still the most important goal of long-tailed learning in the future.

By now, SADE~\cite{zhang2021test} achieves the best overall performance in terms of accuracy and RA (c.f. Table~\ref{table_performance}), but SADE does not perform state-of-the-art on all class subsets (c.f. Table~\ref{table_setperformance}). For example, when training 200 epochs, the head-class performance of SADE is worse than RIDE and its tail-class performance is worse than BBN. To summarize, the higher average performance of SADE implies that the key to obtaining better long-tailed performance is a better trade-off among all classes. In summary, the current best practice for deep long-tailed learning  is using ensemble learning and class re-balancing, simultaneously.


 \begin{table}[t] 
	\caption{{Results on iNaturalist 2018  in terms of accuracy  under 200 training epochs. In this table, CR, IA and MI indicate  class re-balancing, information augmentation and  module improvement, respectively.}} 
	\vspace{-0.1in}
	\label{table_inat} 
 \begin{center}
 \begin{threeparttable}  
    \resizebox{0.4\textwidth}{!}{ 
     	\begin{tabular}{llcccc}\toprule  
        Type & Method      & Head  & Middle & Tail &  All \cr
        \midrule
         Baseline & Softmax	&\textbf{75.3}& 66.4& 60.4& 64.9   \\\midrule
        \multirow{5}{*}{CR}  &Weighted Softmax& 66.5&68.0&69.2&68.3  \\  
        
        & Focal loss~\cite{lin2017focal}& 58.8& 66.5& 66.8&  66.6   \\
         &LDAM~\cite{cao2019learning} & 57.4 & 62.7 & 63.8 & 62.8 \\ 
         &Balanced Softmax~\cite{jiawei2020balanced}& 70.9 & 70.7& 70.4& 70.6  \\
         & LADE~\cite{hong2020disentangling}  & 70.1 & 69.5 & 69.9 & 69.7   \\
        \midrule

       \multirow{2}{*}{IA} & SSP~\cite{yang2020rethinking}& 72.0&68.9&66.3&68.2    \\
        & RSG~\cite{wang2021rsg}   & 70.7  & 69.9  &  69.3  & 70.0  \\
        \midrule
       
        \multirow{7}{*}{MI}  
        & PaCo~\cite{cui2021parametric} & 68.5 & 72.0  & 71.8 & 71.6   \\ 
        & Decouple-IB-CRT~\cite{kang2019decoupling}  &  73.2& 68.8& 65.1& 67.8   \\
        &Decouple-IB-LWS~\cite{kang2019decoupling}  & 71.3 & 69.2 & 68.1& 69.0   \\ 
        &MiSLAS~\cite{zhong2021improving}  & 71.7   &  71.5   &  69.7   &  70.7  \\  
         
        &ResLT~\cite{cui2021reslt}& 67.5&  69.2&  70.1&  69.4    \\
        &RIDE~\cite{wang2020long} &71.5  & 70.0   & 71.6  & 71.8  \\  
        & SADE~\cite{zhang2021test}& 74.4  & \textbf{72.5}  & \textbf{73.1}   & \textbf{72.9}    \\ 

        \bottomrule

	\end{tabular}}
	 \end{threeparttable}
	 \end{center}\vspace{-0.15in}
 \end{table}
 
     


\subsection{{Results on  iNaturalist 2018}}

{iNaturalist 2018 is not a  synthetic dataset sampled from a larger data pool, so we cannot build a corresponding \emph{balanced training set  with a similar data size} for it through sampling.  As a result, it is infeasible to compute relative accuracy for it, so we only report   the  performance   in terms of accuracy. As shown in Table~\ref{table_inat}, most observations are similar to those on ImageNet-LT. For example, most long-tailed  methods outperform Softmax. Although LDAM    (based on the official codes)  performs slightly worse, its tail-class performance is better than the baseline,  which demonstrates  that LDAM can alleviate the class imbalance. However, its head-class performance drops significantly due to the head-tail trade-off, thus leading to poor overall performance. In addition, the current state-of-the-art method is SADE~\cite{zhang2021test}   in terms of accuracy, which further demonstrates the superiority of   ensemble-based methods over other types of methods. All these baselines, except  data augmentation based methods,  adopt only basic augmentation operations. If we adopt  stronger data augmentation and longer training,  their model performance can be further improved.}

 \begin{table}[t] 
\caption{{Analysis of class re-balancing on ImageNet-LT based on ResNeXt-50.  LA indicates logit post-adjustment, while re-sampling indicates class-balance re-sampling~\cite{kang2019decoupling}.  BS indicates   Balanced Softmax~\cite{jiawei2020balanced}.}} 
\vspace{-0.1in}
\label{table_rebalancing_analysis} 
\begin{center}
\begin{threeparttable}  
\resizebox{0.38\textwidth}{!}{ 
 	\begin{tabular}{lcccccc}\toprule  
    Loss &  LA    & Re-sampling &    Head  & Middle & Tail &  All \cr
    \midrule
     Softmax   & \tikzxmark  & \tikzxmark  	&66.9 & 40.4&  12.6&  46.8    \\\midrule
     
     \multirow{4}{*}{BS~\cite{jiawei2020balanced}}  & \tikzxmark  & \tikzxmark  	&62.4 & 47.7 & 32.1 & 51.2  \\
     & \tikzcmark  & \tikzxmark  	& 47.2	& 45.5	& 48.5	& 46.6    \\
      & \tikzxmark  & \tikzcmark  	&  57.6	&47.5	&30.6	&49.1 \\
     & \tikzcmark  & \tikzcmark  	& 42.6	&46.6	&43.6	&44.6   \\  
    \bottomrule 
\end{tabular}}
 \end{threeparttable}
 \end{center}\vspace{-0.15in}
\end{table}

 \begin{table}[t] 
	\caption{{Analysis of whether transfer-based methods (\eg SSP pre-training~\cite{yang2020rethinking}) are beneficial to other types of long-tailed learning.  Here, we use ResNet-50 as the backbone since SSP provides an open-source self-supervised pre-trained ResNet-50.}} 
	\vspace{-0.1in}
	\label{table_pretraining_analysis} 
 \begin{center}
 \begin{threeparttable}  
    \resizebox{0.4\textwidth}{!}{ 
     	\begin{tabular}{lccccc}\toprule  
        Method & SSP pre-training~\cite{yang2020rethinking}      & Head  & Middle & Tail &  All \cr
        \midrule
         \multirow{1}{*}{Softmax}  & \tikzxmark   	&   64.7	& 35.9	& 7.1	&43.1  \\ 
                 \midrule
         \multirow{2}{*}{Re-sampling~\cite{kang2019decoupling}}  & \tikzxmark   	&  51.7	& 48.2	& 32.4	& 47.4  \\
         & \tikzcmark  &  63.5& 45.3& 20.5& 49.0 \\ 
                 \midrule
         \multirow{2}{*}{BS~\cite{jiawei2020balanced}}  & \tikzxmark   	&61.7& 47.8& 28.5& 50.5       \\
         & \tikzcmark  &  62.9	& 50.0	& 30.4	& 52.3  \\ 
                 \midrule 
         \multirow{2}{*}{Decouple~\cite{kang2019decoupling}}  & \tikzxmark   	& 64.2	& 46.1	& 26.0	& 50.3   \\
         & \tikzcmark  & 67.3	& 49.1	& 28.3	& 53.3  \\ 
                 \midrule
         \multirow{2}{*}{SADE~\cite{zhang2021test}}  & \tikzxmark   	& 66.0& 56.1& 41.0& 57.8     \\ 
         & \tikzcmark  & 66.3	&56.9	&42.4	& 58.6   \\ 
        \bottomrule 
	\end{tabular}}
	 \end{threeparttable}
	 \end{center}\vspace{-0.15in}
 \end{table}

\subsection{Analysis}

We next analyze the relationship between various types of methods.

\textbf{{Discussions on class re-balancing.}}
{Class re-balancing has three subtypes of methods, \ie re-sampling, class-sensitive learning and logit adjustment. Although they have  the same goal for re-balancing classes,  they are  exclusive to each other to some degree.  As shown in Table~\ref{table_rebalancing_analysis},  Balanced Softmax (class-sensitive learning) alone greatly outperforms Softmax. However, when further using logit adjustment, it performs only comparably to Softmax. 
The reason is that the trained model by class-sensitive learning is already relatively class-balanced, so further using logit adjustment  to post-adjust model inference will  cause the predictions to become biased again and result  in inferior performance. 
The performance is even worse when further combining class-balanced re-sampling. Therefore, simply combining existing class re-balancing without a careful design cannot lead to better performance.}


 
\textbf{{Discussions on the relationship between pre-training and other long-tailed methods.}}
{As mentioned in Section~\ref{sec_IA}, model pre-training is a    transfer-based scheme for long-tailed learning. In this experiment, we  analyze whether it is beneficial to other long-tailed paradigms. As shown in Table~\ref{table_pretraining_analysis}, SSP   pre-training brings consistent performance gains to class re-balancing (class-balanced sampling~\cite{kang2019decoupling} and BS~\cite{jiawei2020balanced}) and module improvement  (Decouple~\cite{kang2019decoupling} and SADE~\cite{zhang2021test}). We thus conclude that transfer-based  methods are complementary to other    long-tailed   paradigms.}

\textbf{{Discussions on the relationship between data augmentation and other long-tailed methods.}}
{We then analyze whether data augmentation methods are beneficial to other long-tailed paradigms. As shown in Table~\ref{table_augment_analysis},  RandAugment~\cite{cubuk2020randaugment}   brings consistent performance improvement to BS (a class re-balancing method), PaCo (representation learning), De-confound (classifier design) and SADE  (ensemble learning). Such a result demonstrates that augmentation-based  methods are complementary to other paradigms of   long-tailed learning.}
 
 \begin{table}[t] 
	\caption{{Analysis of whether augmentation methods (\eg RandAugment) are beneficial to other types of long-tailed learning, based on ResNeXt-50.}} 
	\vspace{-0.1in}
	\label{table_augment_analysis} 
 \begin{center}
 \begin{threeparttable}  
    \resizebox{0.4\textwidth}{!}{ 
     	\begin{tabular}{lccccc}\toprule  
        Method & RandAugment~\cite{cubuk2020randaugment}        & Head  & Middle & Tail &  All \cr
        \midrule
         \multirow{1}{*}{Softmax}  & \tikzxmark   	&66.9 & 40.4&  12.6&  46.8    \\ 
                 \midrule
         \multirow{2}{*}{BS~\cite{jiawei2020balanced}}  & \tikzxmark   	&62.4 & 47.7 & 32.1 & 51.2    \\
         & \tikzcmark  & 64.1 & 50.4 & 32.3 & 53.2  \\ 
                 \midrule
         \multirow{2}{*}{PaCo~\cite{cui2021parametric}}  & \tikzxmark   	&63.2	&51.6 	&39.2	&54.4  \\
         & \tikzcmark  & 63.7  & 56.6   & 39.2  & 57.0  \\ 
                 \midrule
         \multirow{2}{*}{De-confound~\cite{tang2020long}}  & \tikzxmark   	&64.9	&46.9	&28.1	&51.3   \\
         & \tikzcmark  & 66.1  &50.5  & 32.2  & 54.0 \\ 
                 \midrule
         \multirow{2}{*}{SADE~\cite{zhang2021test}}  & \tikzxmark   	& 67.2  	&  55.3 	& 40.0	&    58.8      \\ 
         & \tikzcmark  &67.3	&60.4	&46.4	&61.2  \\ 
        \bottomrule 
	\end{tabular}}
	 \end{threeparttable}
	 \end{center}\vspace{-0.15in}
 \end{table}

 \begin{table}[t] 
	\caption{The decoupled training performance of various class-sensitive losses     under 200 training epochs on ImageNet-LT. Here, ``Joint" indicates one-stage end-to-end joint training; ``NCM" is the nearest class mean classifier~\cite{kang2019decoupling}; ``CRT" represents class-balanced classifier re-training~\cite{kang2019decoupling}; ``LWS" means learnable weight scaling~\cite{kang2019decoupling}. Moreover, BS indicates  the balanced softmax method~\cite{jiawei2020balanced}.} 
	\vspace{-0.1in}
	\label{table_exploration} 
 \begin{center}
 \begin{threeparttable}  
    \resizebox{0.44\textwidth}{!}{
 	\begin{tabular}{lccccccccc}\toprule  
        \multirow{4}{*}{Test Dist.}  &\multicolumn{4}{c}{Accuracy on \textbf{all} classes}&&\multicolumn{4}{c}{Accuracy on \textbf{head} classes}\cr\cmidrule{2-5}\cmidrule{7-10}
        & Joint  & NCM & CRT &  LWS && Joint  & NCM & CRT &  LWS \cr
        \midrule
        Softmax  & 46.8 	&  50.2 	& 50.2  & 50.8  &&  66.9	& 63.5  &  65.0 & 64.6  \\ 
        Focal loss~\cite{lin2017focal} & 47.2	 	& 50.7 & 50.7  & 51.5  &&  67.0 	&  62.6  & 64.5 & 64.3  \\
        ESQL~\cite{tan2020equalization}  & 48.0 	 	&  49.8 & 50.6  & 50.5 &&  63.1	& 60.2   & 64.0 & 63.3 \\
        BS~\cite{jiawei2020balanced}    &  51.2	 	& 50.4	& 50.6	& 51.1 &&  62.4	& 62.4  & 64.9 & 64.3\\ \midrule \midrule

        \multirow{4}{*}{Test Dist.}  &\multicolumn{4}{c}{Accuracy on \textbf{middle} classes}&&\multicolumn{4}{c}{Accuracy on \textbf{tail} classes}\cr\cmidrule{2-5}\cmidrule{7-10}
        & Joint  & NCM & CRT &  LWS && Joint  & NCM & CRT &  LWS \cr
        \midrule
        Softmax   	&  40.4	& 45.8  & 45.3  & 46.1 &&  12.6	& 28.1  & 25.5 & 28.2  \\ 
        Focal loss~\cite{lin2017focal} & 41.0	 	&  47.0	& 46.4 & 47.3  &&  13.1 	& 30.1   & 26.9 & 30.2  \\
        ESQL~\cite{tan2020equalization}  & 44.6	 	&  46.6  & 46.5  &46.1  &&  17.2 	& 31.1  & 27.1 & 29.5  \\
        BS~\cite{jiawei2020balanced}    &  47.7	 	& 46.8 	& 46.1 	& 46.7 && 32.1	& 29.1   & 26.2 & 29.4 \\

        \bottomrule

	\end{tabular}}
	 \end{threeparttable}
	 \end{center}\vspace{-0.15in}
 \end{table}

\textbf{Discussions on class-sensitive losses in the decoupled training scheme.}
We further evaluate the performance of different class-sensitive learning losses     on the decoupled training scheme~\cite{kang2019decoupling}. In the first stage, we use different class-sensitive learning losses to train the model backbone for learning representations, while in the second stage, we use four different strategies for    classifier training~\cite{kang2019decoupling}, \ie  joint training without re-training, the nearest class mean classifier (NCM),  class-balanced classifier re-training (CRT),  and learnable weight scaling (LWS). {As shown in Table~\ref{table_exploration}, decoupled training can further improve  the   overall performance  of most class-sensitive methods with joint training, except  BS. Among these methods, BS   performs the best  under joint training, but the others   perform  comparably to BS  under decoupled training. Such results are particularly interesting, as they imply that although these class-sensitive losses perform differently under joint training, they essentially learn  the similar  quality of feature representations. The worse overall performance of BS under decoupled training than joint training may imply that  BS has conducted class re-balancing very well; further using classifier re-training for   re-balancing does not bring additional benefits but even degenerates the consistency of network parameters  by end-to-end joint training.}

\subsection{{Summary of Empirical Observations}}
{We then  summarize    main take-home messages from our empirical studies.} 
 {First, we analyze to what extent existing long-tailed methods resolve the class imbalance in terms of  relative accuracy, and confirm that   existing research is   making positive progress in resolving class imbalance instead of  just chasing state-of-the-art performance through tricks.}  {Second, we determine the relative performance of existing long-tailed   methods in a unified setup, and find that   ensemble-based methods are the current state-of-the-art.}   {Third, we analyze   method performance on various class subsets, and find that most  methods improve tail-class performance  at the cost of  lower head-class performance. Considering that all classes are important in long-tailed learning, it is worth exploring how to improve all classes at the same time in the future.}
  {Fourth, we empirically show that the three subtypes of class re-balancing are exclusive to each other to some degree. Moreover, information augmentation   methods are complementary to other long-tailed  paradigms.}  {Lastly, by evaluating class-sensitive learning  on the decoupled training scheme, we find   class re-balancing and decoupled training play an interchangeable role in resolving class imbalance. Moreover, the representations learned by different class-sensitive   losses perform similarly under decoupled training.}

\section{{Future Directions}}\label{Sec6}
{In this section, we identify several future research directions for deep long-tailed learning.}

\textbf{Test-agnostic long-tailed learning.} 
Existing long-tailed learning methods generally hypothesize a balanced test class distribution. The practical test distribution, however, often violates this hypothesis (\eg being long-tailed  or even inversely long-tailed), which may lead existing methods to fail in real-world applications.  To overcome this limitation, LADE~\cite{hong2020disentangling} relaxes this hypothesis by assuming that the test class distribution can be skewed arbitrarily but the prior of test distribution is available. Afterward, SADE~\cite{zhang2021test} further innovates the task, in which the test class distribution is not only arbitrarily skewed  but also unknown.  Besides  class imbalance, this task  poses another challenge, \ie unidentified class distribution shift between the training and test samples.  

\textbf{Open-set long-tailed learning.} 
Real-world samples often have a long-tailed and open-ended class distribution. Open-set long-tailed learning~\cite{liu2019large,zhu2020inflated} seeks to learn from long-tailed data and optimize the classification accuracy over a balanced test set that includes head, tail and open classes. There are two main challenges: (1) how to share visual  knowledge between head and tail classes; (2) how to reduce confusion between  tail and open classes.

\textbf{Federated long-tailed learning.}
Existing long-tailed   studies generally assume that all   training samples are accessible during model training. However, in real  applications, long-tailed training data may be distributed on numerous mobile devices or the Internet of Things~\cite{luo2021no}, which requires decentralized training of deep models. Such a task   is called   federated long-tailed learning,  which has  two key challenges: (1) long-tail class imbalance; (2) unknown class distribution shift among the local data of different clients.

\textbf{Class-incremental long-tailed learning.}
In real-world applications, long-tailed data may come in a continual and class-incremental manner~\cite{kim2020imbalanced,hu2020learning,niu2021adaxpert}. To deal with this scenario, class-incremental long-tailed learning  aims to learn deep models from class-incremental long-tailed data, suffering two key challenges: (1) how to handle long-tailed class imbalance when different classes come sequentially, and the model has no information about the future input regarding classes as well as label frequencies; (2) how to overcome catastrophic forgetting of previous class knowledge when learning new classes.  Such a task setting can also be named continual long-tailed learning.

\textbf{Multi-domain  long-tailed learning.} Current long-tailed  methods generally assume that all long-tailed samples come from the same data marginal distribution. However, in practice, long-tailed data may  also get from different domains with distinct data distributions~\cite{zhang2020covid,jamal2020rethinking}, \eg the DomainNet dataset~\cite{peng2019moment}. Motivated by this, multi-domain long-tailed learning  seeks to  handle both class imbalance and domain distribution shift, simultaneously. One more challenging issue may be the inconsistency of class imbalance among different domains. In other words, various domains may have different class distributions, which further enlarges the domain   shift in multi-domain long-tailed learning.

\textbf{Robust long-tailed learning.}
Real-world  long-tailed samples may also suffer image noise~\cite{wu2021adversarial,cao2020heteroskedastic} or label noise~\cite{wei2021robust,karthik2021learning}.  Most   long-tailed methods, however, assume all images and labels are clean, leading to poor model robustness in practical applications. This issue would be  particularly severe for tail classes, as they have very limited training samples.   Inspired by this, robust long-tailed learning seeks to  handle  the class imbalance and improve model robustness, simultaneously.

\textbf{Long-tailed regression.}
Most existing studies of long-tailed visual learning focus on classification, detection and segmentation, which have discrete labels with class indices. However, many tasks involve continuous labels, where hard classification boundaries among classes do not exist. Motivated by this, long-tailed regression~\cite{yang2021delving} aims to deal with   long-tailed learning with continuous label space. In such a task, how to simultaneously resolve long-tailed class imbalance and handle potential missing data for certain labels  remains an open question.

\textbf{Long-tailed video learning.}
Most existing deep long-tailed learning studies focus on the image level, but ignore that the video domain also suffers from the issue of long-tail class imbalance. 
Considering the additional temporal dimension in video data, long-tailed video learning should be more difficult than long-tailed image learning. 
Thanks to the recent release of a  VideoLT dataset~\cite{zhang2021videolt}, long-tailed video learning can be explored in the near future.

\section{Conclusion}\label{Sec7}
In this survey, we have extensively  reviewed classic deep long-tailed  learning methods proposed before mid-2021, according to the taxonomy of class re-balancing, information augmentation and module improvement. We have   empirically   analyzed several state-of-the-art long-tailed methods by evaluating to what extent they   address the issue of class imbalance, based on a newly proposed relative accuracy metric. Following that, we discussed the main application scenarios of long-tailed learning, and identified   potential innovation directions for methods  and   task settings.  We expect that this timely survey not only provides a better understanding of long-tailed learning for researchers and the community, but also facilitates future research.

\section*{Acknowledgements}
This work was partially supported by NUS ODPRT Grant A-0008067-00-00.
  
{
\bibliographystyle{IEEEtran}
\bibliography{LT_survey}
}

         

      

\begin{IEEEbiography}[{\includegraphics[width=1.1in,height=1.3in,clip,keepaspectratio]{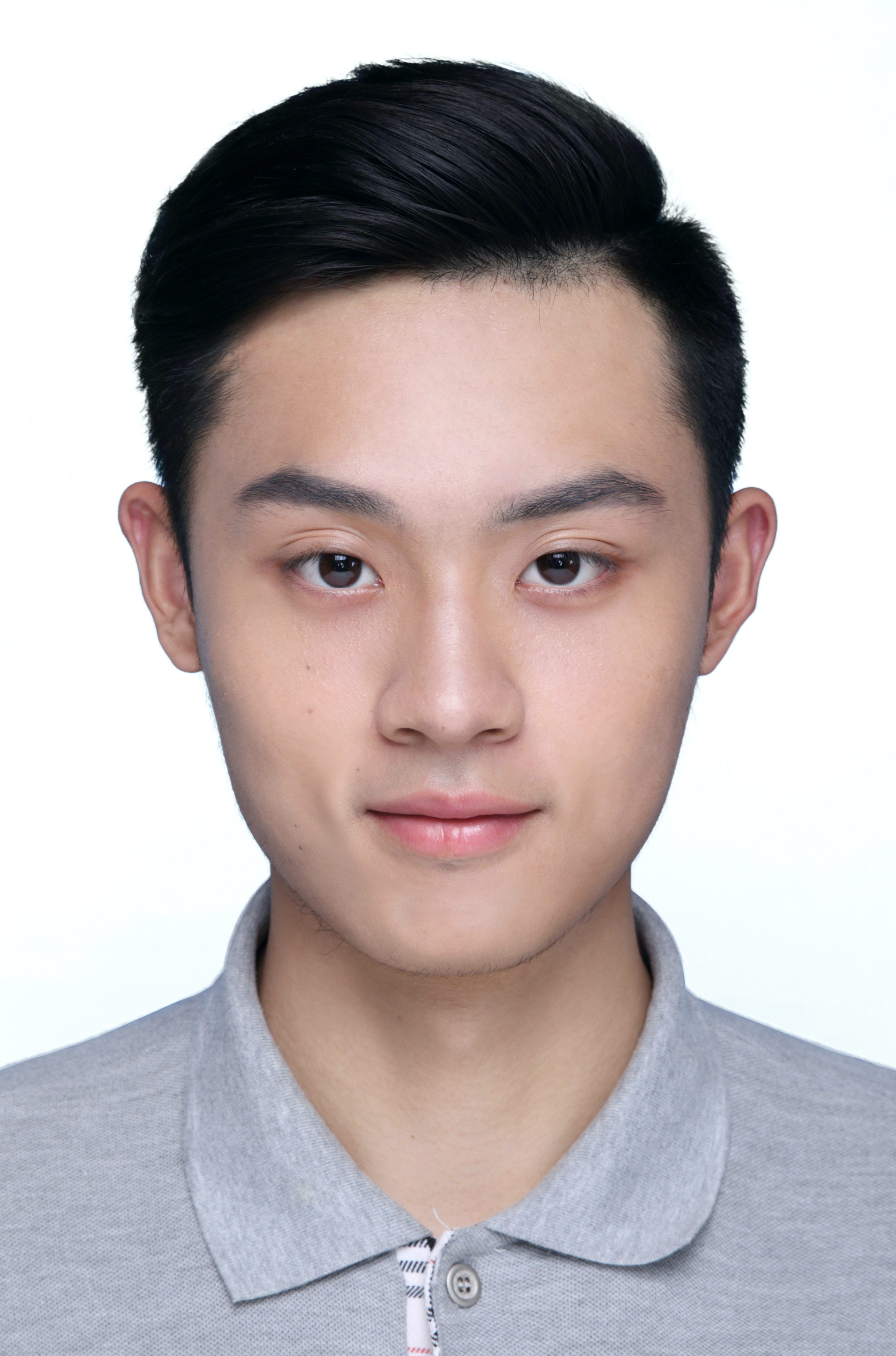}}]{Yifan Zhang}
 is working toward the Ph.D. degree in computer science at National University of Singapore. His research interests are broadly in machine learning, now with high self-motivation to solve domain shifts problems for deep learning. He has published papers in top venues, including NeurIPS, ICML, ICLR, SIGKDD, ECCV, IJCAI, TPAMI, TIP, and TKDE. He has been invited as a  reviewer for top-tier conferences and journals, including NeurIPS, ICML, ICLR, CVPR, ECCV, AAAI, IJCAI, TPAMI, TIP, IJCV, and JMLR.
\end{IEEEbiography}

\begin{IEEEbiography}[{\includegraphics[width=1in,height=1.25in,clip]{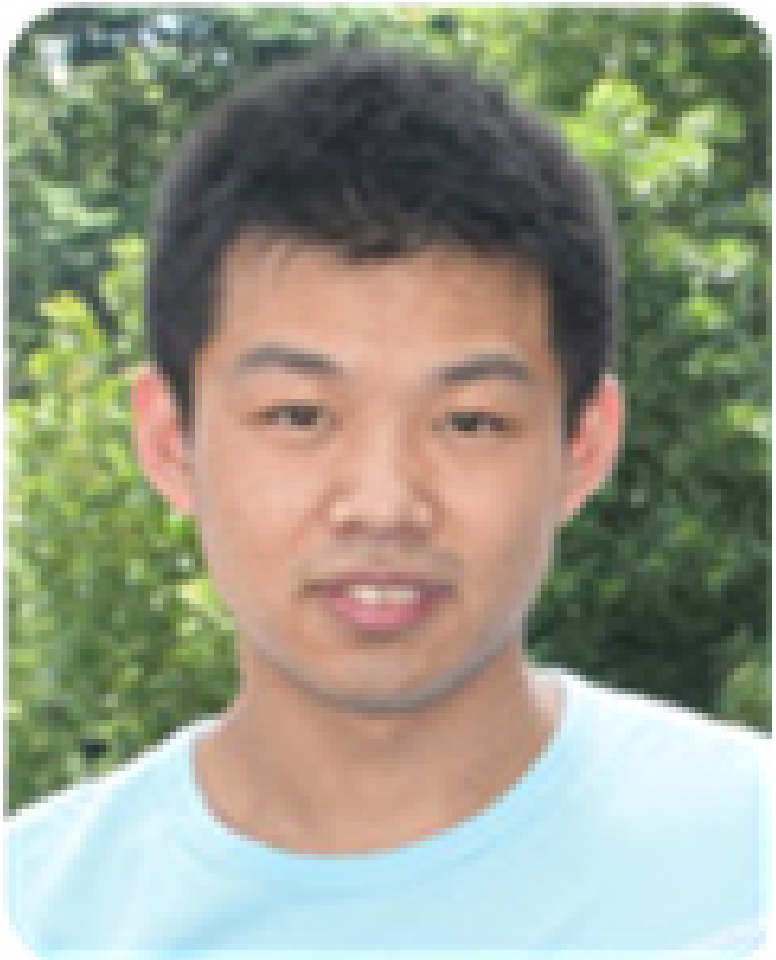}}]{Bingyi Kang} is currently a research scientist at TikTok. Before joining TikTok, got his Ph.D degree in Electronic and Computer Engineering from National University of Singapore. He  received his B.E. degree in automation from Zhejiang University, Hangzhou,
Zhejiang in 2016.  His current research interest focuses on sample-efficient learning and  reinforcement learning.
\end{IEEEbiography}

\begin{IEEEbiography}[{\includegraphics[width=1in,height=1.25in,clip,keepaspectratio]{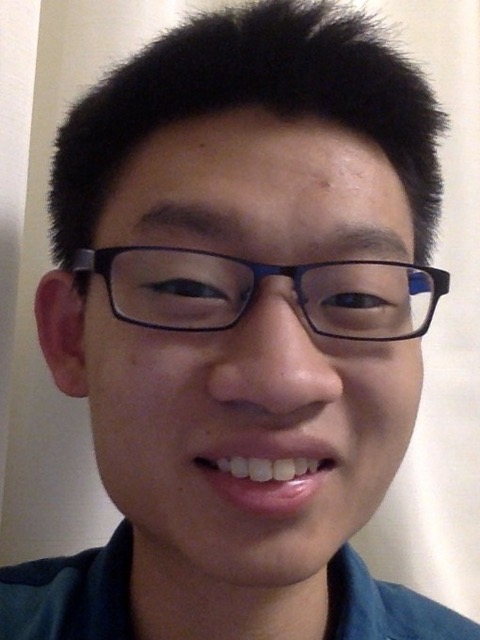}}]{Bryan Hooi} is an assistant professor in the School of Computing and the Institute of Data Science in National University of Singapore. He received his PhD degree in Machine Learning from Carnegie Mellon University, USA in 2019. His research interests include methods for learning from graphs and other complex or multimodal datasets, with the goal of developing efficient and practical approaches for applications such as the detection of anomalies or malicious behavior, and automatic monitoring of medical, traffic, and environmental sensor data. 
\end{IEEEbiography}

\begin{IEEEbiography}[{\includegraphics[width=1in,height=1.4in,clip,keepaspectratio]{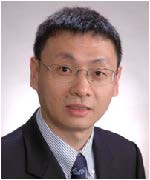}}]{Shuicheng Yan} is currently the director of Sea AI Lab  and group chief scientist of Sea. He is an IEEE Fellow, ACM Fellow, IAPR Fellow, and Fellow of Academy of Engineering, Singapore. His research areas include computer vision, machine learning and multimedia analysis. Till now, he has published over 1,000 papers in top international journals and conferences, with Google Scholar Citation over 93,000 times and H-index 137. He had been among “Thomson Reuters Highly Cited Researchers” in 2014, 2015, 2016, 2018, 2019. His team has received winner or honorable-mention prizes for 10 times of two core competitions, Pascal VOC and ImageNet (ILSVRC), which are deemed as “World Cup” in the computer vision community. Also, his team won over 10 best paper or best student paper prizes and especially, a grand slam in ACM MM, the top conference in multimedia, including Best Paper Award, Best Student Paper Award and Best Demo Award.
\end{IEEEbiography}  

\begin{IEEEbiography}[{\includegraphics[width=1in,height=1.4in,clip,keepaspectratio]{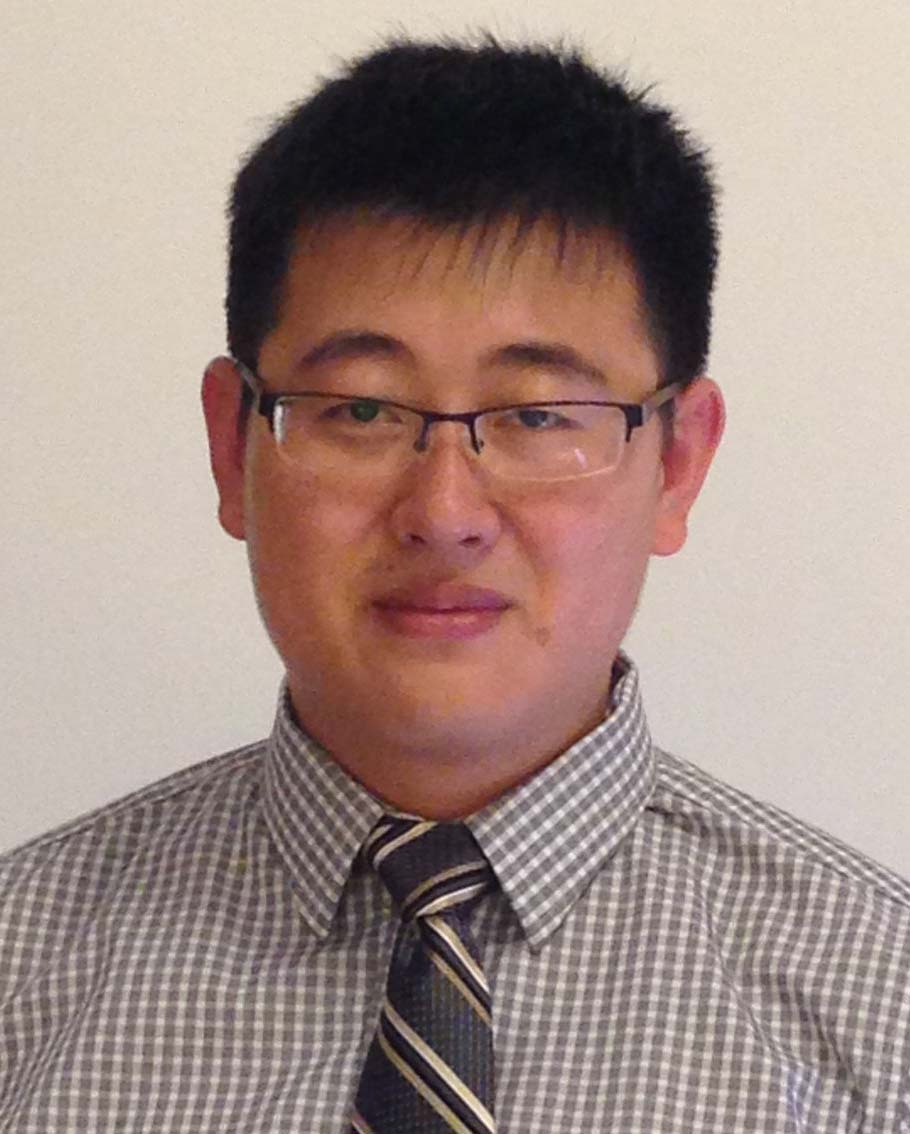}}]{Jiashi Feng} is currently a research manager at TikTok and is leading a fundamental research team. Before  TikTok, he was an assistant professor with  Department of Electrical and Computer Engineering at National University of Singapore and a postdoc researcher in the EECS department and ICSI at the University of California, Berkeley. He received his Ph.D. degree from NUS in 2014.  His research areas include deep learning and their applications in computer vision. He has authored/co-authored more than 300 technical papers on deep learning, image classification, object detection, machine learning theory. His recent research interest focuses on foundation models, transfer learning, 3D vision and deep neural networks. He received the best technical demo award from ACM MM 2012, best paper award from TASK-CV ICCV 2015, best student paper award from ACM MM 2018. He is also the recipient of Innovators Under 35 Asia, MIT Technology Review 2018. He served as the area chairs for NeurIPS, ICML, CVPR, ICLR, WACV, ACM MM and program chair for ICMR 2017.
\end{IEEEbiography} 

\end{document}